\documentclass[conference]{IEEEtran}
\IEEEoverridecommandlockouts
\usepackage{cite}
\usepackage{amsmath,amssymb,amsfonts}
\usepackage{graphicx}
\usepackage{textcomp}
\usepackage{xcolor}

\usepackage[flushleft]{threeparttable}
\usepackage{tabularx}
\renewcommand{\arraystretch}{1.2}
\usepackage{amsthm}
\usepackage{multirow}
\usepackage{algorithm}
\usepackage{algorithmicx}
\usepackage{algpseudocode}
\usepackage{caption}
\usepackage{subcaption}
\usepackage{mathtools}
\usepackage{booktabs}
\usepackage{footnote}
\usepackage{url}
\DeclarePairedDelimiter{\ceil}{\lceil}{\rceil}

\def\BibTeX{{\rm B\kern-.05em{\sc i\kern-.025em b}\kern-.08em
    T\kern-.1667em\lower.7ex\hbox{E}\kern-.125emX}}
\begin{document}

\title{Position-based Hash Embeddings For Scaling \\Graph Neural Networks\\
%
}

\author{\IEEEauthorblockN{Maria Kalantzi}
\IEEEauthorblockA{\textit{Computer Science \& Engineering} \\
\textit{University of Minnesota}\\
Minneapolis, USA \\
kalan028@umn.edu}
\and
\IEEEauthorblockN{George Karypis}
\IEEEauthorblockA{\textit{Computer Science \& Engineering} \\
\textit{University of Minnesota}\\
Minneapolis, USA \\
karypis@umn.edu}
}

\maketitle

\begin{abstract}
Graph Neural Networks (GNNs) bring the power of deep representation learning to graph and relational data and achieve state-of-the-art performance in many applications. GNNs compute node representations by taking into account the topology of the node's ego-network and the features of the ego-network's nodes. When the nodes do not have high-quality features, GNNs learn an embedding layer to compute node embeddings and use them as input features. However, the size of the embedding layer is linear to the product of the number of nodes in the graph and the dimensionality of the embedding
and does not scale to big data and graphs with hundreds of millions of nodes. To reduce the memory associated with this embedding layer, hashing-based approaches, commonly used in applications like NLP and recommender systems, can potentially be used. However, a direct application of these ideas fails to exploit the fact that in many real-world graphs, nodes that are topologically close will tend to be related to each other (homophily) and as such their representations will be similar.

In this work, we present approaches that take advantage of the nodes' position in the graph to dramatically reduce the memory required, with minimal if any degradation in the quality of the resulting GNN model. Our approaches decompose a node's embedding into two components: a \textit{position-specific} component and a \textit{node-specific} component. The position-specific component models homophily and the node-specific component models the node-to-node variation. Extensive experiments using different datasets and GNN models show that our methods are able to reduce the memory requirements by $88\%$ to $97\%$ while achieving, in nearly all cases, better classification accuracy than other competing approaches, including the full embeddings. 
\end{abstract}


\section{Introduction}
In recent years graph neural networks (GNNs) have seen great success and have been widely applied to problems from computer vision~\cite{te2018rgcnn} and natural language processing (NLP)~\cite{marcheggiani2018exploiting, manchanda2021importance} to chemistry~\cite{shui2020heterogeneous} and recommender systems~\cite{ying2018graph}.
%
GNNs compute node representations by considering both the topology of the graph and the nodes' features in end-to-end training. When there are no features available or not enough features for training the model, GNNs use node identity features (one-hot encodings) and learn an embedding layer to compute node embeddings. Then they use these embeddings as input features.
%
%
%
%
Learning an embedding layer for the one-hot node features leads to an embedding table whose size is equal to the product of the number of nodes in the graph and the dimensionality of the embedding. This induces memory requirements that grow linearly with the size of the embedding table and for large graphs with hundreds of millions of nodes, the embedding dimension can range from hundreds to thousands of dimensions.
%
%

In domains such as NLP~\cite{weinberger2009feature, svenstrup2017hash} and recommender systems~\cite{kang2020deep,serra2017getting,zhang2020model}, hashing-based techniques have been developed to reduce the size of the embedding table.
The hashing trick~\cite{weinberger2009feature} uses a hash function to randomly maps the IDs of the features to a smaller number of shared embeddings (hash buckets). However, this method suffers from collisions, as multiple IDs are mapped to the same bucket. 
Other methods build upon the hashing trick and reduce collisions by using multiple hash functions~\cite{serra2017getting, zhang2020model} and learnable feature-dedicated importance weights~\cite{svenstrup2017hash}. A recent work, DHE~\cite{kang2020deep} replaces one-hot encodings with dense hash encodings and trains a deep feedforward network to get the final embeddings. This method reports state-of-the-art performance compared to other hash-based techniques.

Even though GNNs can take advantage of the above hashing-based methods, most real-world graphs have certain properties that can be exploited to develop better methods. 
One such property is network homophily, according to which similar nodes based on node attributes, more likely may attach to each other than dissimilar ones~\cite{mcpherson2001birds}. In most real-world graphs, this leads to nodes that are topologically close in the graph tend to have similar representations.
%

In this work, we propose a family of methods for position-based node embedding learning. Our methods offer a memory efficient alternative to the expensive embedding table coming from the one-hot node features. 
%
The final embeddings consist of two components: a \textit{position-specific} component and a \textit{node-specific} component. 
The first is designed to capture the position of a node in the graph's topology and exploit the fact that nodes which have similar positions (be close together in the graph or share the same set of neighbors) will most likely have similar embeddings due to homophily.
The second is designed to model node-to-node variation and more localized signals.
 

We developed two different approaches for computing the \textit{position-specific} component. 
In the first one, we perform a $k$-way graph partitioning to identify $k$ partitions of nodes. Then, we learn a unique embedding for each partition. All the nodes that belong to the same partition are assigned the same embedding: the embedding of their partition. 
%
In the second one, we build a hierarchy of partitions. 
A partition higher in the hierarchy captures the relations and interactions of the partitions in the previous level. 
We learn a unique embedding for each partition at each level. The final embedding of a node is the combination of the embeddings of the partitions it belongs to, along its hierarchical path. 

For the computation of the \textit{node-specific} component, we developed two approaches based on Hash Embeddings~\cite{svenstrup2017hash}. 
%
%
In the first one, we take into account the hierarchy of partitions. We distribute the embeddings equally among the partitions of the highest hierarchical level and nodes that belong to the same partition share a certain number of embeddings.
The second one does not account for the hierarchy. The embeddings are shared among all nodes, irrespective of the partition they belong to.
We evaluate our methods on three benchmark graph datasets provided by Open Graph Benchmark~\cite{hu2020open} for the task of node property prediction. For each dataset, we test two different state-of-the-art GNN models. As our experimental results showed, our methods perform better than existing approaches, including the one-hot full embeddings, in nearly all cases, and at the same time, they reduce the amount of memory required to compute the initial node embeddings by $88\%$ up to $97\%$ for the largest dataset considered.

We summarize our main contributions in the following:
\begin{itemize}
    \item We introduce hashing to GNN node embedding learning for reducing the amount of memory required to compute the initial node embeddings. We evaluate the performance of various state-of-the-art hashing-based methods in settings where one-hot encodings are used.
    \item We present a family of methods for position-based node embedding learning. These methods offer a wide spectrum of model compression and can be used accordingly.
    We explore experimentally the importance of each method, as well as the benefit we receive in performance. 
    \item We show empirically that our methods achieve high quality memory reduction by leveraging the nodes' position in the graph. This constitutes them suitable and scalable to even extremely large graphs. We reduce the number of trainable parameters by $88\%$ to $97\%$ across the different datasets and models. 
    \item We compare our methods against state-of-the-art hashing-based approaches and show that our methods perform better in almost all cases.
    \item The fact that our methods lead to better performance compared to full embeddings, and at the same time reduce dramatically the number of trainable parameters, indicates that we do not need the full capacity of full embeddings to get high quality node representations.  
\end{itemize}

In Section~\ref{sec:notation} we discuss one-hot full embeddings in the context of GNNs and the basics from hashing-based embeddings. In Sections~\ref{PosEmb} and~\ref{experiments}, we present our methods and extensive experimental evaluation. We continue with discussing related work in Section~\ref{rw} and in Section~\ref{conclusion}, we conclude our work by summarizing the main points.

\begin{figure}[!t]
\centerline{\includegraphics[width=0.9\linewidth]{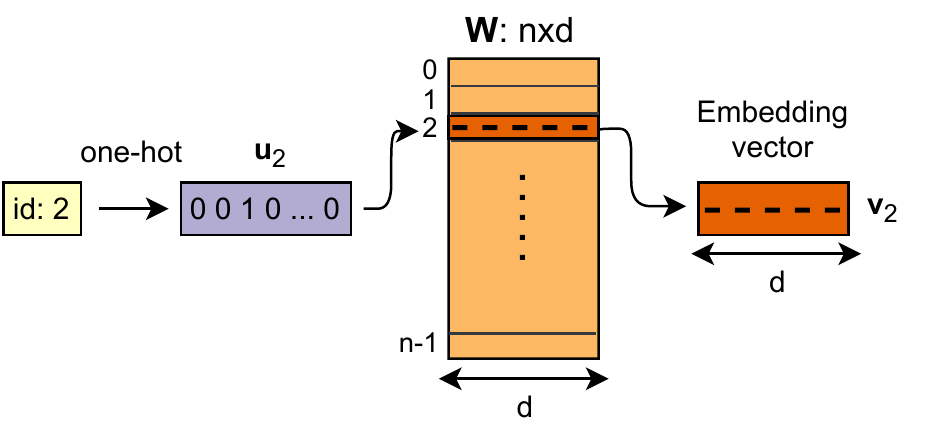}}
\caption{Illustration of one-hot full embeddings for example node with id $2$. The one-hot encoding for this node, vector $u_2$ of dimension $n$, maps to the corresponding entry in the embedding table, $\mathbf{W}$ of size $n\times d$, which contains the final embedding of the node, $\mathbf{v}_2$, a vector of dimension $d$.}
\label{fig:fullembeds}
\end{figure}

\section{Notation and Preliminaries}
\label{sec:notation}
In this section we discuss GNNs and one-hot full embeddings as well as hashing-based embeddings. The notation used throughout the paper is summarized in Table~\ref{table:notation}. 

\subsection{GNNs and One-hot Full Embeddings}
Graph Neural Networks (GNNs) are deep learning models that operate on graph structured data. They stack GNN layers to extract topological signals and learn node embeddings.
Most of the layers in a GNN model can be expressed under a framework of message passing. Each node sends/receives messages to/from its neighbor nodes. A message is a function of three things: (i) the embedding of the source node, (ii) the embedding of the destination node and (iii) the edge features, if available.
The message passing framework includes two phases: message passing and node update. Each node $i$ updates its embedding by receiving messages from its neighbors:
\begin{equation}
\label{message_passing}
\begin{aligned}
    \mathbf{m}_i^{(t)} &= \sum_{j \in \mathcal{N}(i)} f^{(t)} \Big( \mathbf{h}_i^{(t)}, \mathbf{h}_j^{(t)}, \mathbf{e}_{ij} \Big) \\
    &\mathbf{h}_i^{(t+1)} = g^{(t)} \Big( \mathbf{h}_i^{(t)}, \mathbf{m}_i^{(t)} \Big),
\end{aligned}
\end{equation}
where $\mathcal{N}(i)$ is the set of neighbor nodes of $i$, $\mathbf{h}_i^{(t)}$ is the node embedding of $i$ in layer $t$, $\mathbf{m}_i^{(t)}$ is a message vector created by aggregating messages from $i$’s neighbor nodes, $\mathbf{e}_{ij}$ is the edge feature associated with the edge between $i$ and $j$, $f(\cdot)$ is a learnable function that maps embeddings of the sender and receiver as well as the corresponding edge feature to a message vector, $g(\cdot)$ is a learnable function that updates the node embedding by combining the incoming message and the embedding from the previous layer. 

In cases where there are no input features or these are of low-quality, the most straightforward way for GNNs to proceed is to use one-hot encodings and learn node embeddings as the input features. This leads to the full embedding table $\mathbf{W} \in \mathbb{R}^{n\times d}$. 
Let $\mathbf{v}_i$ be the embedding of node $i$. Then, we have
\begin{equation}
    \mathbf{v}_i = \mathbf{W}^T \mathbf{u}_i,
\end{equation}
where $\mathbf{u}_i$ is the one-hot encoding vector for node $i$, i.e., $\mathbf{u}_i \in \{0,1\}^n$, with $u_i(i)=1$ and $u_i(t)=0$, $\forall t \neq i$, and $u_i(t)$ is the $t$-th component of vector $\mathbf{u}_i$. 
In Figure~\ref{fig:fullembeds} we can see an illustration of the full embeddings for a node example.

Then, in Eq.~\ref{message_passing}, we have
\begin{equation}
    \mathbf{h}_i^{(0)}=\mathbf{v}_i.
\end{equation}
The size of the embedding table in this case is $n\times d$, which is not scalable for graphs with hundreds of millions or billions of nodes and for which cases $d$ may range from hundreds to thousands of dimensions.

\subsection{Hashing-based Embeddings}
Hashing-based techniques reduce the size of the embedding table by using hash functions to map feature values to shared learnable embeddings (hash buckets). 
Let $B \ll n$ be the number of hash buckets, i.e., the number of rows of the embedding table $\mathbf{W}$. 
The hashing trick~\cite{weinberger2009feature} uses a single hash function and distributes the hashed values uniformly. The main drawback of this method is that it suffers from collisions as $B\ll n$. In order to reduce collisions, other methods use multiple hash functions~\cite{serra2017getting, svenstrup2017hash, zhang2020model}.

Let $\mathbf{u}_i \in \{0,1\}^{B}$ be the vector we get after applying a hash function $H$ to node $i$, with $i \in \{0, \dots, n-1\}$. The hash function maps $i$ to $\{0,1,\dots,B-1\}$ and $u_i (s)=1$ when $s = H(i)$, and $u_i (t)=0$, $\forall t \neq s$. 
When $h$ hash functions are used, then $\mathbf{u}_i$ consists of $h$ component vectors, $\mathbf{u}_i = [\mathbf{u}_i^{(1)}; \mathbf{u}_i^{(2)}; \dots; \mathbf{u}_i^{(h)}] \in \{0,1\}^{B \times h}$ generated by the $h$ hash functions. The final embedding is the combination of the component vectors depending on the method.
Let $\mathbf{W} \in \mathbb{R}^{B\times d}$ be the weight matrix of the embedding layer. Then, for the basic hashing-based methods, we have:
\begin{itemize}
    \item Hashing trick:
    \begin{equation}
         \mathbf{v}_i = \mathbf{W}^T \mathbf{u}_i.
    \end{equation}
    The size of the embedding table in this case is $B \times d$.
    \item Double hashing: 
    \begin{equation}
        \mathbf{v}_i = \mathbf{W}^T(\mathbf{u}^{(1)}_i+\mathbf{u}_i^{(2)}).
    \end{equation}
    The size of the embedding table in this case is again $B\times d$.
    \item Hash embeddings: 
    \begin{equation}
    \label{eq:hashembeddings}
        \mathbf{v}_i = \mathbf{W}^T(y^{(1)}_i \mathbf{u}_i^{(1)} + y^{(2)}_i \mathbf{u}_i^{(2)}+\dots + y^{(h)}_i \mathbf{u}_i^{(h)}),
    \end{equation}
        where $\mathbf{y}_i$ is the node-specific importance vector of node $i$, which controls the contribution of each of the $h$ component vectors. The size of the embedding table in this case is $B\times d + n\times h$.
\end{itemize}



\begin{table} 
	\centering
	\caption{Notation.}
	\begin{tabular}{ p{0.3\linewidth} p{0.6\linewidth}} 
	    \toprule
	    Notation & Description \\
	    \midrule
		$n \in \mathbb{N}$, & number of nodes\\
		$B \in \mathbb{N}$, & number of hash buckets\\
		$b \in \mathbb{N}$, & number of shared embeddings dedicated to the node-specific term\\
		$\alpha\in \mathbb{N}$, & hyperparameter that controls the number of partitions\\
		$m \in \mathbb{N}$, & total number of partitions\\
		$m_i \in \mathbb{N}$, & number of partitions at level $i$\\
		$d \in \mathbb{N}$, & embedding dimension\\
		$L \in \mathbb{N}$, & the number of hierarchical levels\\
		\mbox{$H:\mathcal{N} \rightarrow [0,\dots,b-1]$}, & hash function for mapping nodes to $b$ embeddings\\
		$h \in \mathbb{N}$, & number of hash functions\\
		$\mathbf{W}\in \mathbb{R}^{B\times d}$, & trainable embedding table of size $B\times d$\\
		$\mathbf{z}_i \in \mathbb{R}^L$, & partition membership vector for node $i$\\
		$\mathbf{P}_i \in \mathbb{R}^{m_i \times d}$, & embedding table containing the partitions' embeddings of level $i$\\
		$c \in \mathbb{N}$, & compression factor for the node-specific term\\
		$\mathbf{X}_i \in \mathbb{R}^{c \times d}$, & embedding table containing the node-specific embeddings of partition $i$ in the finest level\\
		\bottomrule
	\end{tabular}
	\label{table:notation}
\end{table}

\section{Position-based Hash Embeddings (PosHashEmb)}
\label{PosEmb}
%
Graph learning algorithms introduce relational inductive bias, which may potentially lead to similar node representations for nodes that are close together in the graph. In this work, we develop a family of methods, PosHashEmb, which 
leverages homophily by learning position-based embeddings. 
%
Instead of learning a single node-specific embedding that comes solely from the one-hot encoding of the node,
PosHashEmb expresses the embedding of each node as the combination of two components: a \textit{position-specific} component and a \textit{node-specific} component. The position-specific term is designed to model homophily and the node-specific term models the node-distinct characteristics. 
%

Let $\mathbf{p}_i$ be the position-specific component of node $i$ and $\mathbf{x}_i$ the node-specific component.
PosHashEmb computes the final embedding for node $i$, $\mathbf{v}_i$, as the sum of the two components, i.e.,
\begin{equation}
\label{final_emb}
    \mathbf{v}_i = \mathbf{p}_i + \mathbf{x}_i.
\end{equation}
%
%
Following, we describe in detail each component.

\begin{figure*}[!t]
\centerline{\includegraphics[width=0.77\linewidth]{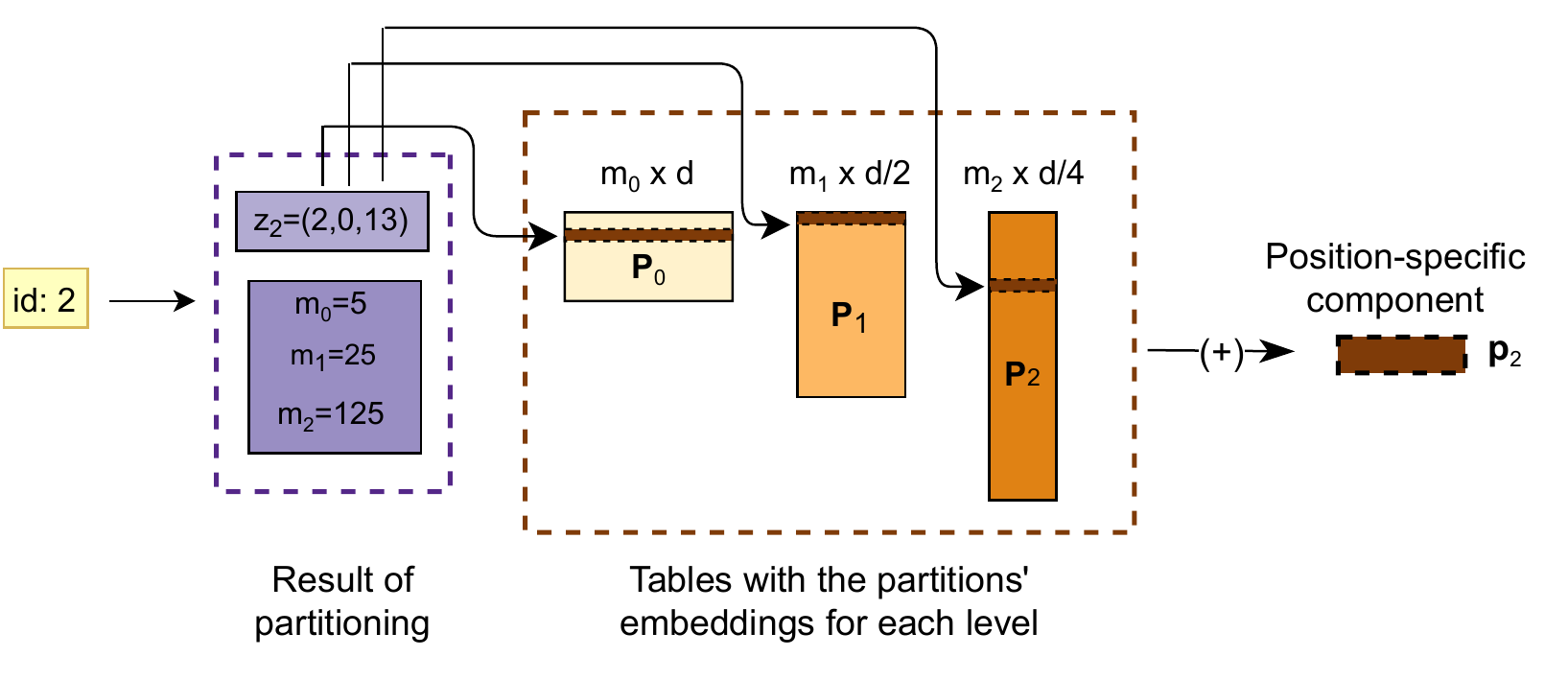}}
\caption{Illustration of the position-specific component for node with id $2$ in the case of multiple hierarchical levels. In this example, we assume the following: $n=625, L=3, k=5$. We perform $5$-way partitioning and we get the membership vectors $\mathbf{z}$ for the nodes. Since $L=3$ and $k=5$ we have $m_0=5, m_1=25, m_2=125$. Then, for computing the position-specific component of node with id $2$, from each level we retrieve the embedding of the partition the node belongs to, according to $\mathbf{z}_2$. Note the different sizes of the three embedding tables. The embeddings of the coarsest level (level 0) are assigned higher embedding dimension (embedding table $\mathbf{P}_0$ with size $m_0 \times d$).  
The position-specific component of the node, $\mathbf{p}_2$ of size $d$, is given by the summation of the three embeddings.}
\label{fig:position_comp}
\end{figure*}
\subsection{Position-specific Component}
\label{sec:topology}
In order to leverage homophily, 
we perform a $k$-way graph partitioning to discover node communities.
The number of partitions $k$, is controlled by a hyperparameter $\alpha$, with $\alpha < 1$, and is given by
\begin{equation}
\label{k}
    k=n^{\alpha}.
\end{equation}
We explore two ways to capture position: either create a single level partitioning consisting of $k$ partitions or a hierarchy of partitions. 

\subsubsection{Single level approach}
We learn a unique embedding for each of the $k$ partitions. The position-specific component of a node embedding is the embedding of the partition it belongs to. In particular, we have $\mathbf{P}\in \mathbb{R}^{k\times d}$ to be the embedding table containing the partitions' embeddings. Then, for node $i$ we have
\begin{equation}
    \mathbf{p}_i = \mathbf{P}[z_i, :],
\end{equation}
where $z_i$ is a scalar indicating the partition id that $i$ belongs to and $\mathbf{P}[z_i, :]$ is the $z_i$-th row of $\mathbf{P}$. 
The size of the embedding table is $k \times d$, where $k \ll n$ and can be extremely small, as we will show in our experiments. We will refer to this method as PosEmb $1$-level.

\subsubsection{Hierarchical approach}
The nodes themselves form the lowest level in the hierarchy, where one partition consists of a single node.
The higher the hierarchical level, the coarser the communities are. 
Each level $i$ in the hierarchy 
captures the relations and interactions of the communities in level $i+1$.
A hierarchy with $L$ levels is constructed by applying recursive $k$-way graph partitioning $L$ times. We assign a number to each level, with the top level numbered $0$, and the bottom level numbered $L-1$.
Level $0$ is obtained by computing a $k$-way partitioning. Level $1$ is obtained by partitioning each of them into $k$ parts, leading to a total of $k^2$ partitions. Subsequent levels are obtained in a similar fashion recursively.
Let $m \in \mathbb{N}$ be the number of total partitions across all levels. We have
\begin{equation}
    m = \sum_{j=0}^{L-1} k^{j+1}.
\end{equation}
Again, we learn a unique embedding for each partition. This corresponds to learning strongly distinguishable representations as we move to higher levels in the hierarchy (coarser representations). 
%
The size of the embedding table then becomes $m \times d$, where $m \ll n$.

Another way to look at the case of multiple levels is to have $L$ embedding tables, one for each hierarchical level. 
We construct each table in such a way that the coarsest level (level $0$) is assigned higher learning capacity for each representation. The motivation behind that is the following. Each partition in this level includes the largest number of nodes which share the same embedding, compared to other levels. This means that there are more data samples available for training. As a result, these embeddings can be estimated more reliably. 
In order to balance the overall model compression, as we move to finer levels, we decrease the embedding dimension.
In particular, we have $\{\mathbf{P}_0, \mathbf{P}_1, \dots, \mathbf{P}_{L-1}\}$, where $\mathbf{P}_i \in \mathbb{R}^{m_i \times d_i}$ is the embedding table containing the partitions' embeddings of level $i$, 
$m_i$ is the number of partitions in level $i$ and $d_i$ is the embedding dimension for level $i$ with $d = d_0 > d_1 > \dots > d_{L-1}$. 
Let $\mathbf{z}_i \in \mathbb{R}^L$ be the membership vector for node $i$ whose $j$-th component, $z_i(j)$, contains the partition id that node $i$ belongs to for level $j$. The vectors $\mathbf{z}$ for all nodes are the output of the partitioning. 
For the final position-specific component of node $i$, we combine the corresponding hierarchical level representations by summing up all the partition embeddings across its hierarchical path. 
We have
\begin{equation}
\label{pi}
    \mathbf{p}_i = \sum_{j=0}^{L-1} \mathbf{P}_j[z_i(j), :],
\end{equation}
where $\mathbf{P}_j[z_i(j), :]$ is the row vector in the $z_i(j)$-th row of table $\mathbf{P}_j$.
Nodes that belong to the same partition share the same position-specific component of the final embedding. Figure~\ref{fig:position_comp} presents an example of computing the position-specific component of a node when there are multiple hierarchical levels.



\subsection{Node-specific Component}
\label{sec:node-specific}
Rather than fitting a unique embedding vector for each node (as in the case of one-hot full embeddings), now each node's embedding is selected from a shared pool of $b$ embeddings ($b$ hash buckets) by using hash functions to map nodes to the buckets.    
We use hashing in order to map the nodes to the shared embeddings and specifically, universal hashing for integers~\cite{carter1979universal}. We follow the hash embeddings~\cite{svenstrup2017hash} paradigm as in Eq.~\ref{eq:hashembeddings}. Specifically, we use $h$ hash functions and for each node, we have $h$ mappings to buckets, which correspond to $h$ component vectors. Then, we learn node-specific importance weights to combine the $h$ component vectors. In particular, the importance parameter vector $\mathbf{y}_i$ for node $i$ is represented as a row in an $n \times h$ trainable matrix $\mathbf{Y}$. Each of the $n$ rows is a vector of $h$ scalar importance parameters. 
$y_i^{(j)}$ is a scalar that corresponds to the importance weight of node $i$ for the $j$-th component vector.
Following, we describe two different ways to distribute the shared embeddings.
\subsubsection{Intra-partition shared embeddings}
The nodes that belong to the same partition in the coarsest level of the hierarchy (level $0$) share $c$ embeddings. There is a total of $b$ shared embeddings and $c=b/m_0$, where $m_0$ is the number of partitions in level 0.
We have $m_0$ embedding tables, $\{\mathbf{X}_0, \mathbf{X}_1, \dots, \mathbf{X}_{m_0-1} \}$, where $\mathbf{X}_i \in \mathbb{R}^{c \times d}$ is the embedding table containing the node-specific embeddings of partition $i$ in level $0$ (the coarsest level) and $c$ is the compression factor and is equal to $b/m_0$. 
%
Then we have
\begin{equation}
\label{xi_intra}
    \mathbf{x}_i = \mathbf{X}_{z_i(0)}^T(y^{(1)}_i \mathbf{u}_i^{(1)} + y^{(2)}_i \mathbf{u}_i^{(2)}+\dots + y^{(h)}_i \mathbf{u}_i^{(h)}).
\end{equation}
This approach can be combined only with the hierarchical approach, as far as the computation of the position-specific component is concerned.

\subsubsection{Inter-partition shared embeddings}
In this approach, following the hash embeddings approach~\cite{svenstrup2017hash}, there is a set of $b$ embeddings that are shared among all the nodes. We learn globally shared node embeddings, where we assign $b$ shared embeddings for the nodes, irrespective of the partition they belong to. 
We have a single embedding table, $\mathbf{X} \in \mathbb{R}^{b \times d}$ and we have
\begin{equation}
\label{xi_inter}
    \mathbf{x}_i = \mathbf{X}^T(y^{(1)}_i \mathbf{u}_i^{(1)} + y^{(2)}_i \mathbf{u}_i^{(2)}+\dots + y^{(h)}_i \mathbf{u}_i^{(h)}).
\end{equation}

The size of the embedding table for the node-specific computation is $b \times d$, where $b \ll n$. We also need some additional space of size $n\times h$ for the importance weights, and typically, $h=1$ or $h=2$.
In Algorithm~\ref{alg:PosEmb}, we describe our method PosHashEmb consisting of both the position-specific and node-specific components described in this section.





\begin{algorithm}[!t] 
	\caption{Position-based Hash Embeddings (PosHashEmb)}
	\label{alg:PosEmb}
	\begin{algorithmic}[1]
		\Require graph $G$, hyperparameter $\alpha$, number of hierarchical levels $L$, embedding dimension $d$, number of nodes $n$, number of hash functions $h$.
		\State $k \gets n^{a}$
		\State $\mathbf{Z}, \mathbf{l} \gets \texttt{metis}(G, k, L)$ \Comment{Call recursive $k$-way metis partitioning. It returns a matrix $\mathbf{Z}$ with rows the membership vectors for all nodes, and a vector $\mathbf{l} \in \mathbb{R}^L$ with values the number of partitions for each level.}
		\State $m_0 \gets l[0]$ \Comment{Number of partitions on level $0$.}
		\State $c \gets \sqrt{\frac{n}{m_0}}$ \Comment{Set the compression factor for the hash buckets of the node-specific term.}
		\State $d' \gets d$
		\For{$i \gets 0$ to $L-1$}
		    \State $\mathbf{P}_i \gets \texttt{Embedding}(m_i,d')$ \Comment{Embedding tables for the position-specific component.}
		    \State $d' \gets d'/2$
		\EndFor
        \For{$i \gets 0$ to $m_0-1$}
            \State $\mathbf{X}_i \gets \texttt{Embedding}(c,d)$ \Comment{Embedding tables for the node-specific component.}
        \EndFor   
        \State $\mathbf{Y} \gets \texttt{Embedding}(n,h)$ \Comment{Trainable matrix for the importance weights of the node-specific component.}
        \State $\lambda \gets 1$
        \For{$epoch$}
        \For{$i \gets 0$ to $n-1$} \Comment{For each node.}
            \State{Compute $\mathbf{p}_i$ according to Eq.~\ref{pi}}
            \State{Compute $\mathbf{x}_i$ according to Eq.~\ref{xi_intra}} or Eq.~\ref{xi_inter}
            \State $\mathbf{v}_i \gets \mathbf{p}_i + \lambda \mathbf{x}_i$
        \EndFor
        \State{$\mathbf{V} \gets \text{stack}(\mathbf{v}_i)$}
		\State \Return $\mathbf{V}$
		\EndFor
	\end{algorithmic}
\end{algorithm}

\section{Experiments}
\label{experiments}
We study the scalability of GNN models on large graphs for the task of node property prediction. We present our extensive experimental evaluation which aims to answer the following research questions:
\begin{itemize}
    \item \textbf{RQ1:} What is the effect of the number of partitions in the performance?
    \item \textbf{RQ2:} Does the position-specific component improve the performance of full embeddings? Is the combination of the two components more beneficial than each of the components alone?
    \item \textbf{RQ3:} How is the performance affected as we increase the number of hierarchical levels?
    \item \textbf{RQ4:} How is the performance affected as we gradually decrease the complexity of the node-specific part of the embedding? 
    \item \textbf{RQ5:} How does our method, PosHashEmb, compare against other hashing-based methods? 
\end{itemize}

\subsection{Datasets}
We use three datasets from Open Graph Benchmark (OGB)~\cite{hu2020open} for the task of node property prediction. The datasets' statistics are presented in Table~\ref{dataset-stats}. We use the default data splits provided by OGB and the same metrics to measure the performance. For ogbn-arxiv and ogbn-products datasets, the prediction task is multi-class classification and accuracy is used as the performance metric.
For the case of ogbn-proteins, the prediction task is multi-label binary classification and the performance is measured by the average of ROC-AUC scores across the different kinds of labels.

\begin{table}
  \caption{Dataset statistics.}
  \label{dataset-stats}
  \centering
  \begin{threeparttable}
  \begin{tabular}{lrrrl}
    \toprule
    Dataset     & \#Nodes     & \#Edges   & \#Tasks   & Metric \\
    \midrule
    ogbn-arxiv  & 169,343 	& 1,166,243 & 1 & Accuracy \\
    ogbn-proteins & 132,534  &	39,561,252 	& 112 & ROC-AUC \\
    ogbn-products & 2,449,029  &	61,859,140 	& 1 & Accuracy\\
    \bottomrule
  \end{tabular}
  \begin{tablenotes}
    \item For the undirected graphs ogbn-proteins and ogbn-products, the loaded graphs will have the doubled number of edges because we add the bidirectional edges automatically.
  \end{tablenotes}
  \end{threeparttable}
\end{table}

\subsection{Baselines}
\label{baselines}
We consider the one-hot full embeddings (FullEmb) to be the method that requires the \textit{full-size} amount of memory for the computation of the initial node embeddings. FullEmb gets as input a one-hot encoding for each node and learns a unique embedding for every node.
%
We compare our method, PosHashEmb, against the following hashing-based approaches:
\begin{itemize}
    \item \textbf{Hashing trick (HashTrick)}~\cite{weinberger2009feature}. This is a traditional method for handling large-vocab categorical features. It uses a single hash function to randomly map feature values into a smaller feature space (hash buckets). 
    \item \textbf{Bloom embeddings (Bloom)}~\cite{serra2017getting}. Inspired by bloom filters~\cite{bloom1970space}, Bloom Embeddings generate a binary encoding by using multiple hash functions. Then an embedding layer is applied to the encoding to retrieve the compact representation for the required feature value.
    \item \textbf{Hash embeddings (HashEmb)}~\cite{svenstrup2017hash}. Hash embeddings use multiple hash functions and retrieve the corresponding entries from the embedding table (component vectors). HashEmb learns feature-specific weights that uses to control the contribution of each component vector for generating the final embeddings. 
    \item \textbf{Deep hash embeddings (DHE)}~\cite{kang2020deep}. This is a recently proposed method with non-one-hot encodings and a deep neural network (DNN) for computing embeddings. DHE first encodes the feature value to a dense vector with multiple hash functions and then applies a DNN to generate the embedding.

\end{itemize}

\subsection{GNN models}
For each dataset, we choose two different GNN models implemented with Deep Graph Library (DGL)~\cite{wang2019dgl}, which perform best based on the OGB leaderboard\footnote{\url{https://ogb.stanford.edu/docs/leader_nodeprop/}}. We use the embedding method on top of the GNN model and we perform end-to-end training. For ogbn-arxiv, we use GCN~\cite{kipf2016semi} and GAT~\cite{velivckovic2017graph} models. For ogbn-proteins, we use MWE-DGCN\footnote{\url{https://cims.nyu.edu/\%7Echenzh/files/GCN_with_edge_weights.pdf}} and GAT~\cite{velivckovic2017graph}. Last, for ogbn-products, we use GRAPHSAGE~\cite{hamilton2017inductive} and GAT~\cite{velivckovic2017graph}. The training parameters for each model are set to those tuned by the DGL team.


\subsection{Implementation details}
\label{sec:impl_details}
We implement all the methods using PyTorch and  DGL~\cite{wang2019dgl}.
We consider the case where we only use the identity features as the input features, i.e., one-hot encodings, and no additional node features, as the former are the ones that lead to large memory requirements increase coming from the size of the embedding table.
 
For the embedding dimension $d$, we used $128$ for ogbn-arxiv and $100$ for ogbn-products; the same as the dimension of the dataset's original node features. For ogbn-proteins, where there were no node features originally, we tested values $\{30,40,50,60,40,80,90,100,150,200,250\}$ and we report results for $d=200$, where the full embeddings performed best. In the case of ogbn-proteins, we kept the $8$-dimensional edge features.
For PosHashEmb, we set $\alpha=1/4$, i.e., $k=n^{1/4}$, $c=\ceil*{\sqrt{n/k}}$ and $b=\ceil*{\sqrt{n/k}} k$, $L=3$ and $d_0=d, d_1=d/2, d_2=d/4$, unless stated otherwise. For the graph partitioning we use \texttt{METIS}~\cite{karypis1997metis}.
%
%
For PosHashEmb, HashEmb and Bloom we use $h=2$.

For DHE, we use $h=1024$ as the number of hash functions used for the computation of the initial dense hash encodings and $B=10^6$ (this does not affect the size of the embedding table); both values proposed by the authors. For the computation of the embeddings, we used the default neural network architecture (equal-width MLP), which is the best performing according to the original paper~\cite{kang2020deep}.
The authors found that embedding networks with around five hidden layers perform better. However, this is not the case for our task, as we observed that a network of that depth performed poorly. To this end, we explored the depth of the network as well as the width of the hidden layers. 
For the number of hidden layers we tried values $\{0,1,2,3,4\}$, for the hidden width size (hidden dimensions) $\{500, 1000, 1500, 2000, 2500,3000\}$, activation functions $\{relu, mish\}$ (the two best performing activation functions proposed by the authors), and we tried every case with or without Batch Normalization~\cite{ioffe2015batch}.
We found small differences in performance regarding the two activation functions and inclusion of batch normalization or not. We report results for the best parameters: one hidden layer, hidden width size of $2000$ and $relu$ activation function.

We run every experiment five times and we report the average performance and standard deviation. For ogbn-arxiv and ogbn-proteins datasets we perform full-batch training. For ogbn-products, we perform mini-batch training; we set the batch size equal to $1000$ and we use as the sampling model the one where each node gathers messages from all its neighbors (full neighbor sampling). We used NVIDIA V-100 GPU for model training and inference. 

\begin{table*}[!t]
  \caption{Performance comparison when we use the position-specific component for the computation of the node embeddings.}
  \label{tab:fullemb-combined}
  \begin{threeparttable}
      \begin{tabularx}{1.0\textwidth}{ 
   >{}X 
   >{\centering\arraybackslash\hsize=.8\hsize}X 
   >{\centering\arraybackslash\hsize=.8\hsize}X 
   >{\centering\arraybackslash\hsize=.8\hsize}X
   >{\centering\arraybackslash\hsize=.8\hsize}X 
   >{\centering\arraybackslash\hsize=.8\hsize}X 
   >{\centering\arraybackslash\hsize=.8\hsize}X }
        \toprule
        & \multicolumn{2}{c}{\textbf{ogbn-arxiv}} & \multicolumn{2}{c}{\textbf{ogbn-products}} & \multicolumn{2}{c}{\textbf{ogbn-proteins}} \\
        Method & GCN     & GAT     & GRAPHSAGE  & GAT & \text{MWE-DGCN} & GAT \\
        \midrule
        FullEmb & \mbox{$0.671 \pm 0.004$} & \mbox{$0.677\pm0.003$}  & \mbox{$0.733\pm0.004$} & \mbox{$0.755\pm0.006$}  & \mbox{$0.745\pm0.019$} & \mbox{$0.752\pm0.002$}   \\
        \mbox{PosEmb $1$-level} & \mbox{$0.673\pm0.003$} & \mbox{$0.670\pm0.005$} & \mbox{$0.760\pm0.003$} & \mbox{$0.762\pm0.008$} & \mbox{$0.772\pm0.021$} & \mbox{$0.800\pm0.006$}  \\
        RandomPart & \mbox{$0.634\pm0.003$} & \mbox{$0.657\pm0.001$} & \mbox{$0.691\pm0.002$} &\mbox{$0.731\pm0.003$} & \mbox{$0.773\pm0.011$} &\mbox{$0.758\pm0.004$}   \\
        PosFullEmb $1$-level & \mbox{$0.678\pm0.002$} & \mbox{$0.674\pm0.003$} & \mbox{$0.751\pm0.004$} & \mbox{$0.760\pm0.002$} &\mbox{$0.766\pm0.014$} & \mbox{$0.761\pm0.006$}  \\
        \bottomrule
      \end{tabularx}
        \begin{tablenotes}
            \item FullEmb is the one-hot full embeddings and we consider its memory requirements as the full size. PosEmb $1$-level is a one-level partitioning and represents a method that accounts solely for the position-specific term. RandomPart method is similar to PosEmb $1$-level, but now, we assign the nodes to partitions in a random way (random partitioning). PosFullEmb $1$-level combines PosEmb $1$-level and FullEmb. 
            The amount of memory required to compute the initial node embedding of RandomPart and PosEmb $1$-level corresponds to approximately $1/12$ of the full size in the cases of ogbn-arxiv and ogbn-proteins, and $1/34$ in the case of ogbn-products. 
            This translates to approximately $90\%$ and $97\%$ memory savings, respectively.
            %
            %
            The memory requirements of PosFullEmb $1$-level is larger than the full size.
            Following OGB, the values for ogbn-arxiv and ogbn-products correspond to accuracy, and for ogbn-proteins to ROC-AUC metric; the higher the values the better. We set $\alpha=1/4$.
        \end{tablenotes}
  \end{threeparttable}
\end{table*}

\subsection{How does $\alpha$ affect the performance? (RQ1)}
\begin{figure}[!t]
    \centering
    \setlength{\tabcolsep}{-0.65em}
    \begin{tabular}{ccc}
     \includegraphics[width=50mm]{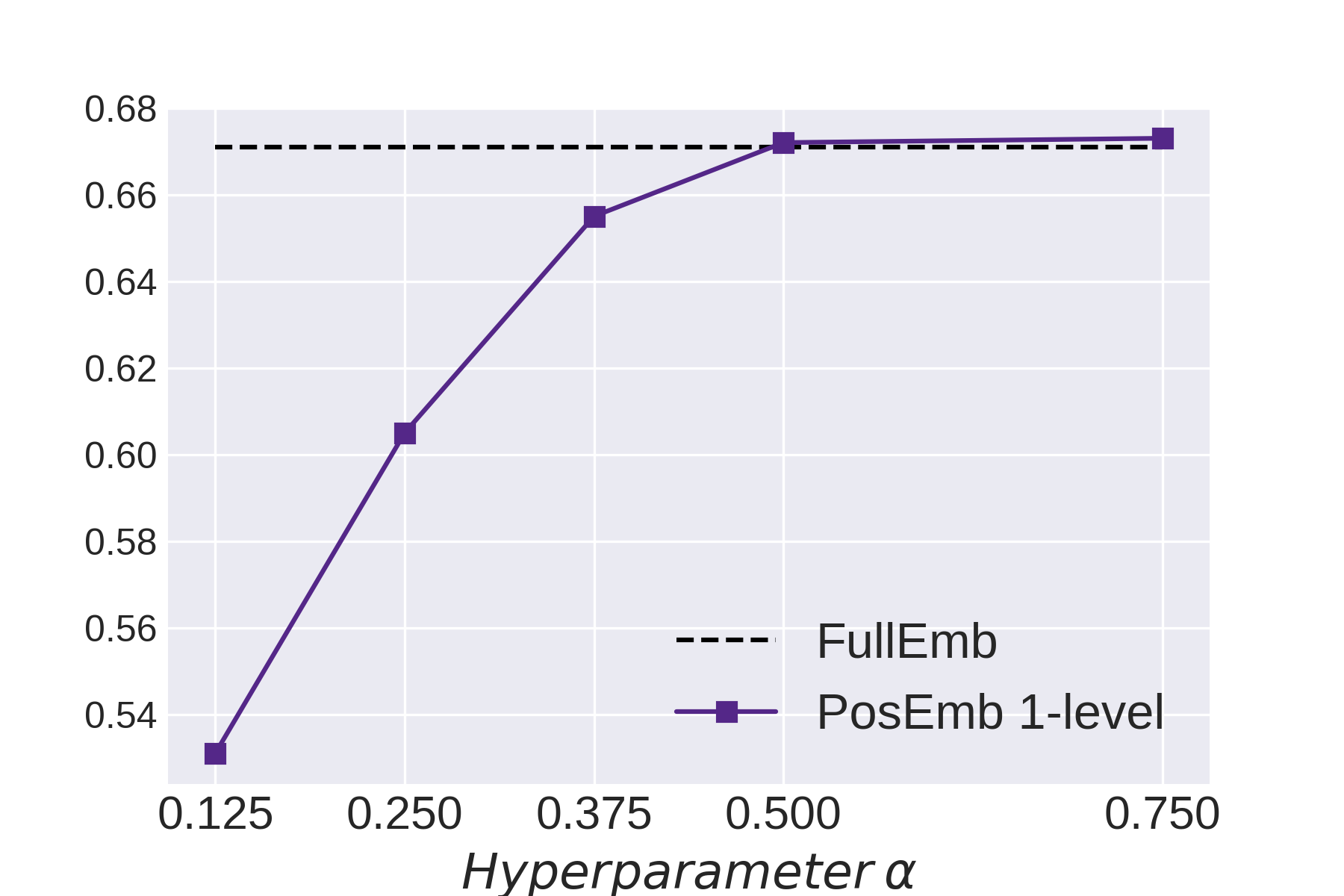} &
     \includegraphics[width=50mm]{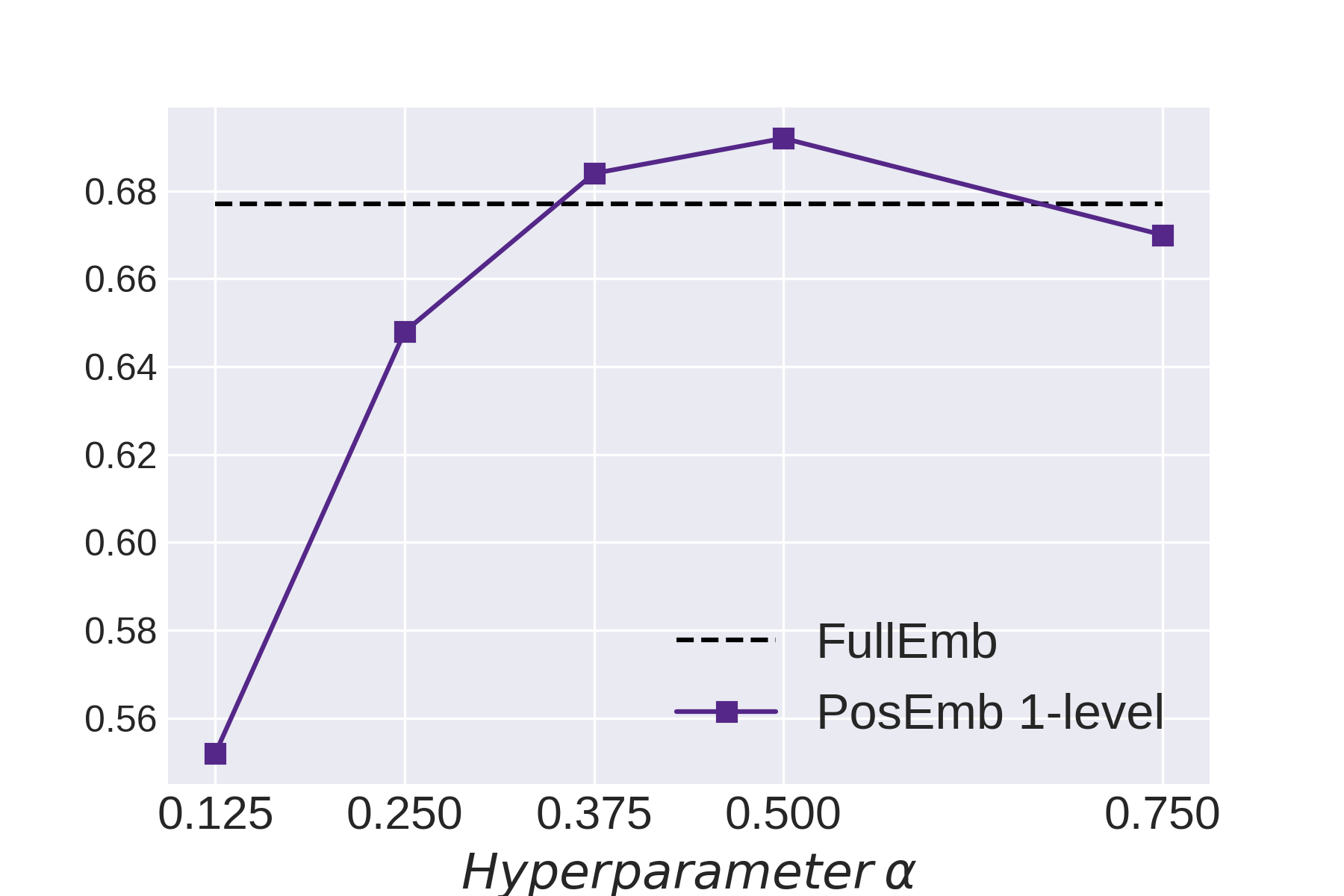} \\
     (a) arxiv, GCN & (b) arxiv, GAT \\
     \includegraphics[width=50mm]{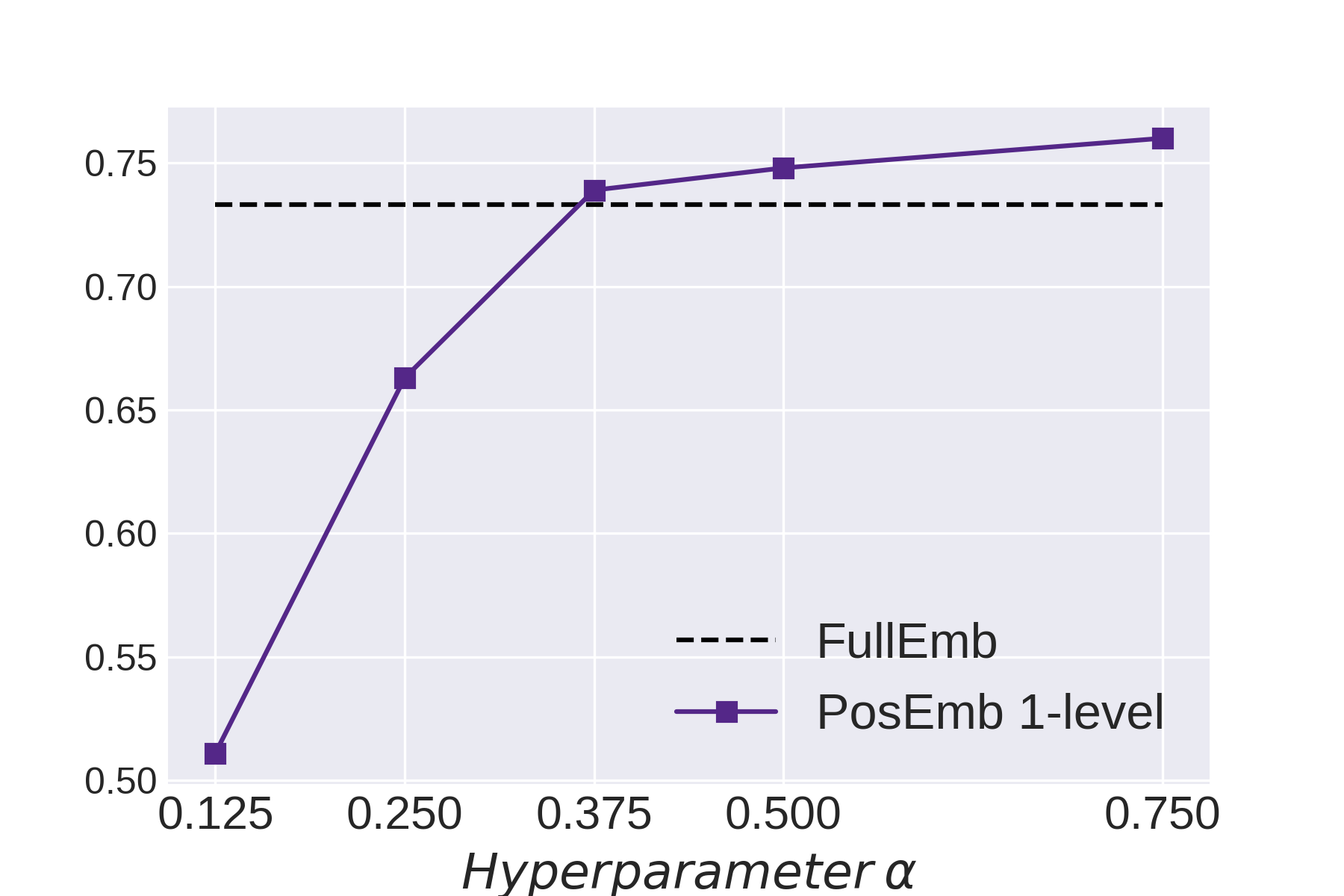} &
     \includegraphics[width=50mm]{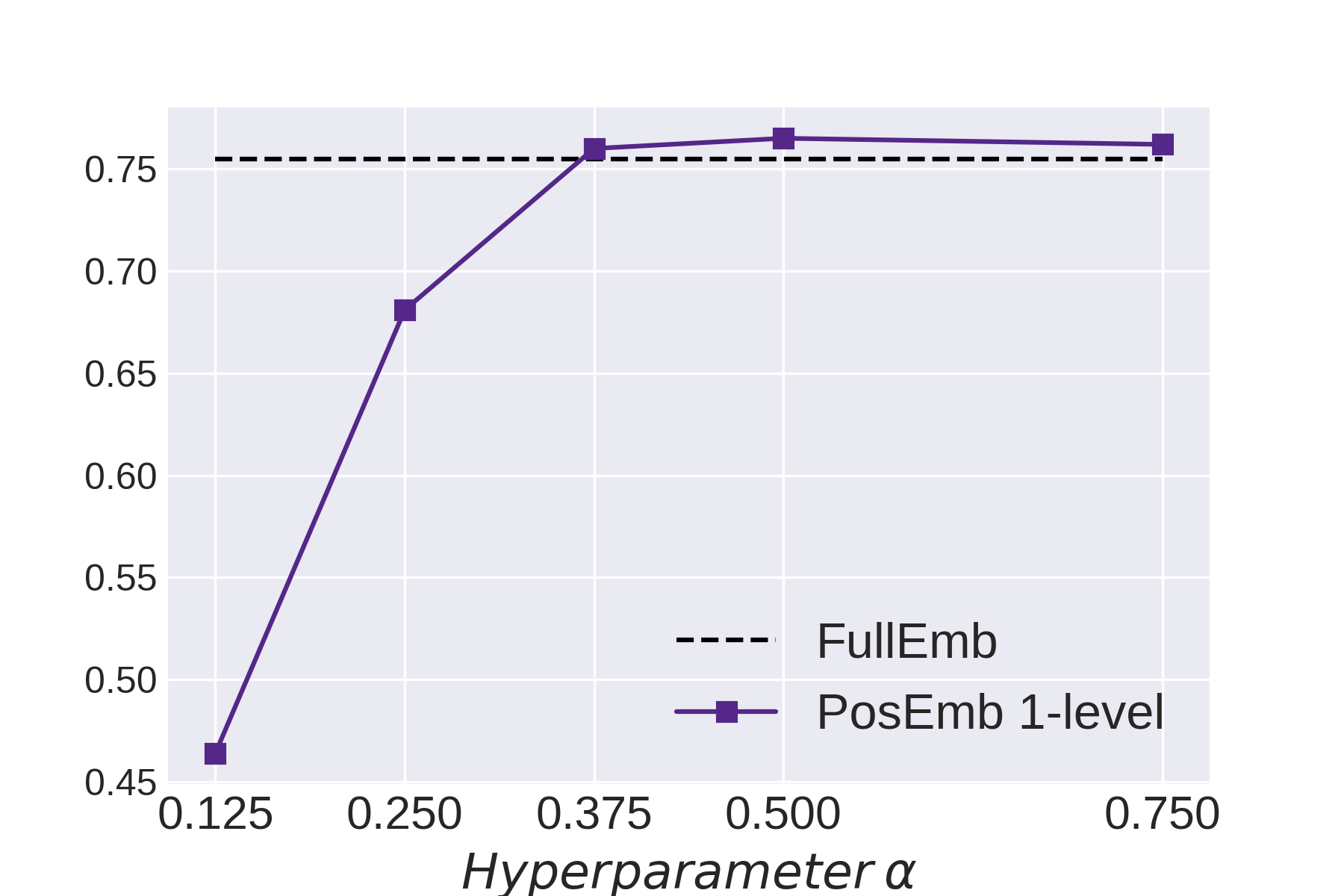} \\
     (c) products, GRAPHSAGE & (d) products, GAT \\
     \includegraphics[width=50mm]{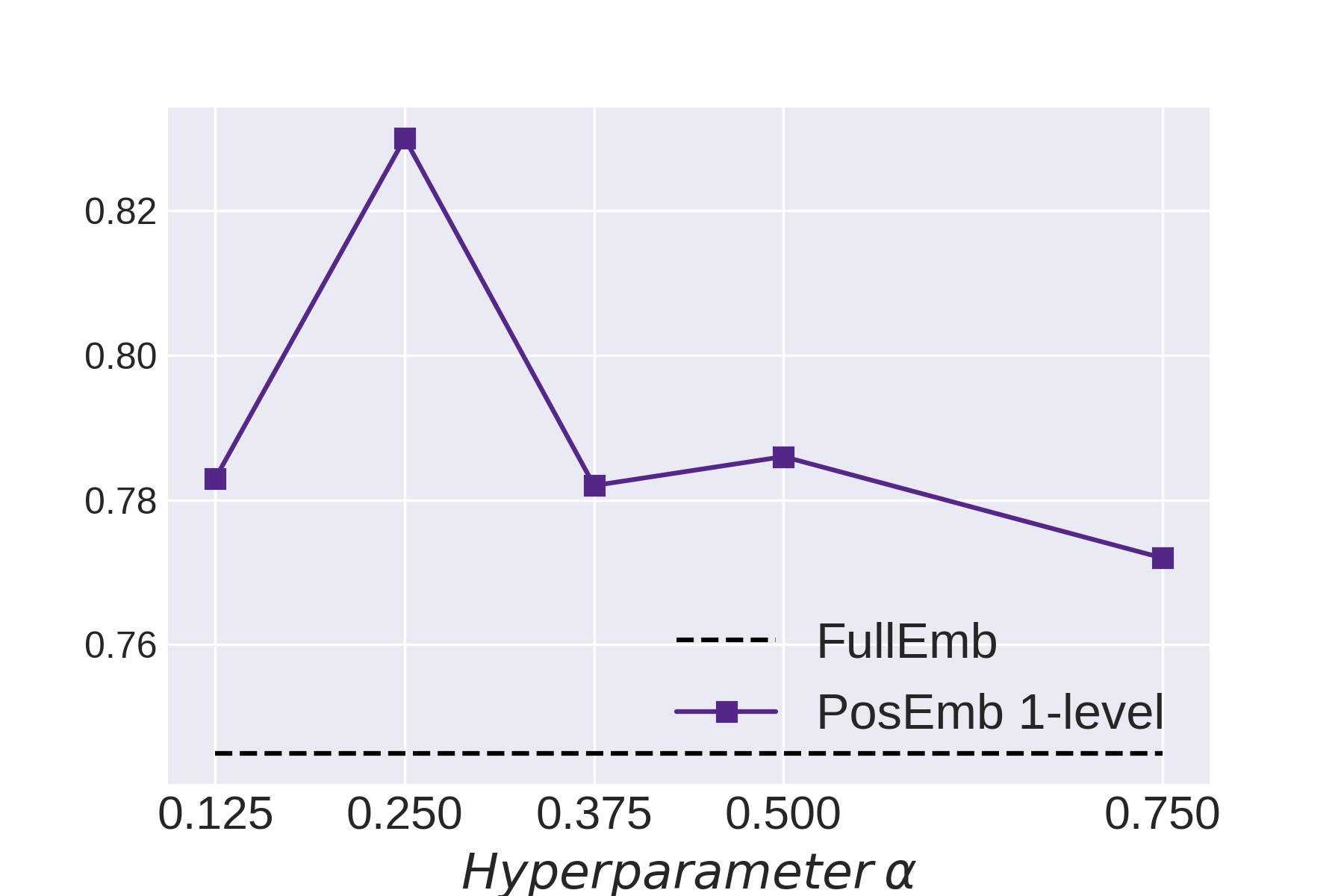} &
     \includegraphics[width=50mm]{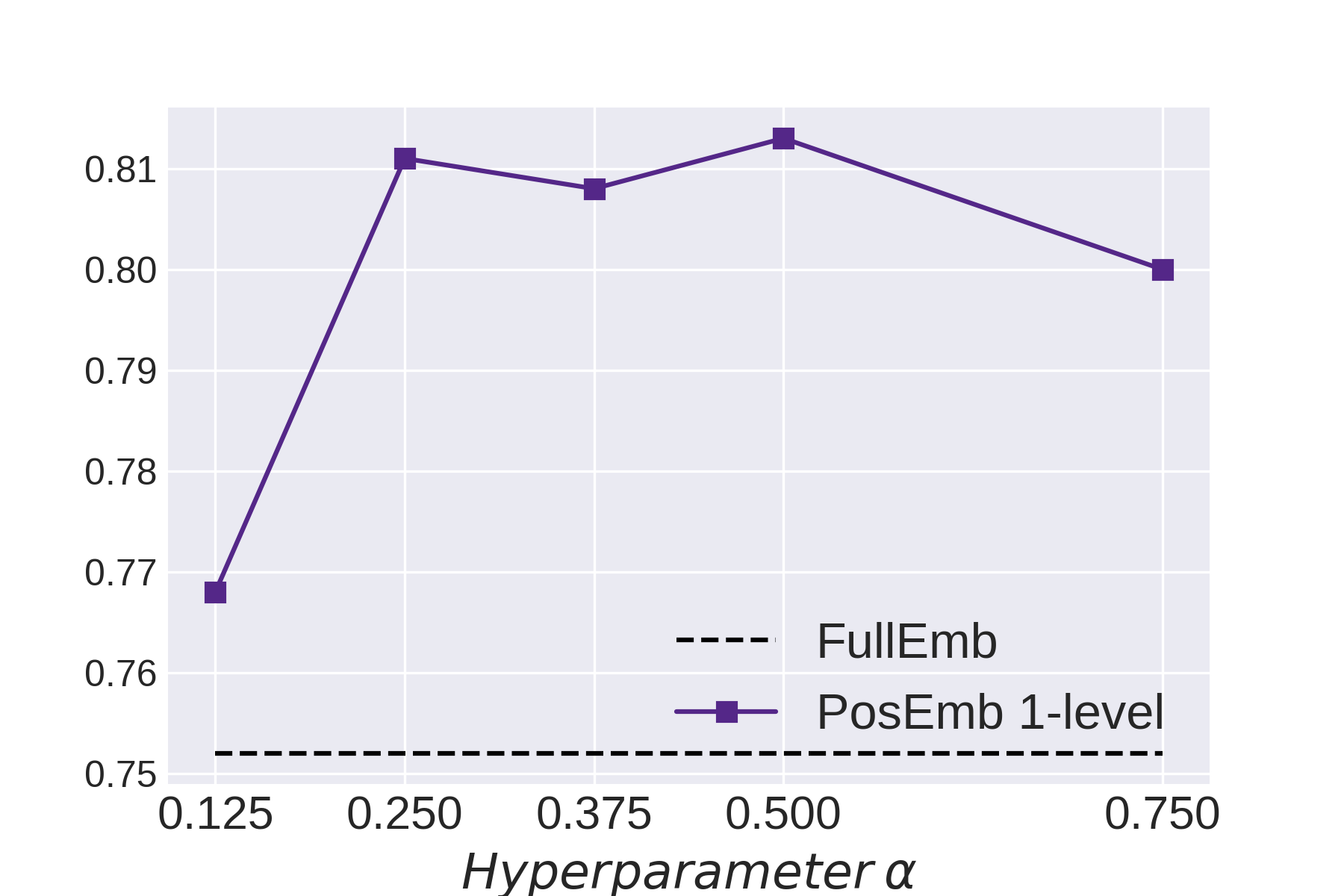} \\
     (e) proteins, MWE-DGCN  & (f) proteins, GAT  \\
    \end{tabular}
    \caption{Performance of PosEmb $1$-level as a function of hyperparameter $\alpha$ for different number of partitions when using a single level partitioning. PosEmb $1$-level consists of the position-specific component solely. Following OGB, for ogbn-arxiv and ogbn-products the performance is measured with accuracy, and for ogbn-proteins with ROC-AUC metric; the higher the values the better.}
    \label{fig:diff_k}
\end{figure}

In order to study how the number of partitions $k$ affect the performance of our method, we only use the position-specific component for the computation of the node embeddings and we use a single level partitioning. We call this method PosEmb $1$-level.
We vary the value of $k$ by controlling hyperparameter $\alpha$. We test the following values for $\alpha$: $\{\frac{1}{8}, \frac{2}{8}, \frac{3}{8}, \frac{4}{8}, \frac{6}{8}\}$. These correspond to the following values for $k$ in each dataset: for ogbn-arxiv $k = \{5,25,125,441,9261\}$, for ogbn-products $k=\{7,40,343,1600,64000\}$ and for ogbn-proteins $k=\{5,25,125,400,8000\}$. We present the results in Figure~\ref{fig:diff_k}. In general, we need a large enough number of partitions to capture the positional relations of the nodes. Up to a certain point, as we increase $k$, either the performance does not significantly change, as in ogbn-products dataset and ogbn-arxiv in the case of GCN, or the performance deteriorates, as in the case of ogbn-proteins for both GNN models and ogbn-arxiv for GAT. One interesting observation is that in the case of ogbn-proteins dataset, even for the smallest number of partitions i.e., $k=5$, the performance is better than full embeddings, and the amount of memory required to compute the initial node embeddings is reduced by $99\%$ for the case of MWE-DGCN and $91\%$ for GAT.

\subsection{Does the position-specific component improve the performance of full embeddings? (RQ2)}
In order to answer this, we start from using the full extent of the node-specific part of the embedding, i.e., the full embeddings (FullEmb), and we combine it with the position-specific term of a single level partitioning (PosEmb $1$-level). We call this method PosFullEmb $1$-level. 
We also compare the performance with PosEmb $1$-level, a method which consists solely by the position-specific component and this comes from a single level partitioning. As we can see in Table~\ref{tab:fullemb-combined}, PosFullEmb $1$-level improves the performance of FullEmb. This means that there is a benefit from exploiting the position of nodes in the graph for embedding learning. We observe that PosEmb $1$-level performs better than FullEmb in all cases, except for GAT on ogbn-arxiv where the performance is slightly worse. In addition, PosEmb $1$-level not only performs better than FullEmb, but it also reduces the memory required to compute the initial node embeddings by $86\%$ for ogbn-proteins and GAT, up to $97\%$ for ogbn-products and GAT. 
Further exploring the gains in performance coming from capturing the nodes' position in the graph, we perform the following experiment. We compare the performance when using a community discovery partitioning (e.g., \texttt{METIS}) versus a random partitioning. In particular, we compare PosEmb $1$-level with a method we call RandomPart that corresponds to the random partitioning. Essentially, RandomPart is a hashing trick with the number of hash buckets $B$ to be equal to the number of partitions $k$.
We present the results in Table~\ref{tab:fullemb-combined} and we focus on the performance of RandomPart and PosEmb $1$-level methods. We can see that PosEmb $1$-level leads to a better performance in all datasets and GNN models.

\begin{table*}[!t]
  \caption{Performance results for the different levels of hierarchy for the computation of the position-specific component of the embedding.}
  \label{tab:hierarchy}
    \setlength{\tabcolsep}{0.7em}
  \begin{threeparttable}
      \begin{tabularx}{1.0\textwidth}{ 
  >{\hsize=.8\hsize}X 
   >{\centering\arraybackslash}X 
   >{\centering\arraybackslash}X 
   >{\centering\arraybackslash}X 
   >{\centering\arraybackslash}X 
   >{\centering\arraybackslash}X 
   >{\centering\arraybackslash}X }
        \toprule
        & \multicolumn{2}{c}{\textbf{ogbn-arxiv}} & \multicolumn{2}{c}{\textbf{ogbn-products}} & \multicolumn{2}{c}{\textbf{ogbn-proteins}} \\
        Method & GCN     & GAT     & GRAPHSAGE  & GAT & \text{MWE-DGCN} & GAT \\
        \midrule
        FullEmb & \mbox{$0.671 \pm 0.004$} & \mbox{$0.677\pm0.003$}  & \mbox{$0.733\pm0.004$} & \mbox{$0.755\pm0.006$}  & \mbox{$0.745\pm0.019$} & \mbox{$0.752\pm0.002$}   \\
        \mbox{PosEmb $1$-level} & \mbox{$0.673\pm0.003$} & \mbox{$0.670\pm0.005$} & \mbox{$0.760\pm0.003$} & \mbox{$0.762\pm0.008$} & \mbox{$0.772\pm0.021$} &\mbox{$0.800\pm0.006$}   \\
        \mbox{PosEmb $2$-level} & \mbox{$0.675 \pm 0.001$} & \mbox{$0.674\pm0.003$} & \mbox{$0.761\pm0.006$} & \mbox{$0.760\pm0.005$} & \mbox{$0.776\pm0.004$} & \mbox{$0.786\pm 0.009$}  \\
        \mbox{PosEmb $3$-level} & \mbox{$0.674 \pm 0.002$} & \mbox{$0.676\pm0.001$} & \mbox{$0.758\pm0.005$} & \mbox{$0.767\pm0.006$} &\mbox{$0.788\pm0.005$} & \mbox{$0.791\pm 0.009$}  \\
        \bottomrule
      \end{tabularx}
        \begin{tablenotes}
            \item The embedding dimension for PosEmb $1$-level is set to $d$, for PosEmb $2$-level is set to $d/2$ and for PosEmb $3$-level to $d/4$. We set $d=128$ for ogbn-arxiv, $d=100$ for ogbn-products and $d=200$ for ogbn-proteins. 
            We also set $\alpha=1/4$. PosEmb consists of the position-specific component solely (irrespective of the number of levels).
            FullEmb corresponds to the one-hot full embeddings; we consider its memory requirements as the full size and serves as a baseline.
            Following OGB, the values for ogbn-arxiv and ogbn-products correspond to accuracy, and for ogbn-proteins to ROC-AUC metric; the higher the values the better. The memory savings coming from PosEmb $3$-level ranges from $90\%$ up to $99\%$ across all datasets and GNN models.
        \end{tablenotes}
  \end{threeparttable}
\end{table*}
\subsection{What is the effect of hierarchy in the performance? (RQ3)}
We focus on the position-specific component and we explore how the performance of our method is affected by including multiple hierarchical levels. We start with a single level of partitioning (PosEmb $1$-level) and we keep on adding levels up to three. We exclude entirely the node-specific component from the embedding computation. The results are presented in Table~\ref{tab:hierarchy}. We observe that as we increase the number of hierarchical levels, the performance either gets better or remains unchanged, except for GAT on ogbn-proteins in which case PosEmb $1$-level performs better. At the same time, PosEmb $3$-level reduces the amount of memory required to compute the initial node embeddings by $90\%$ up to $99\%$ across all datasets and GNN models.


\subsection{How much can we decrease the complexity of the node-specific term? (RQ4)}
To answer this, we use both the components for the computation of the embeddings; we start from the full embeddings and we gradually decrease the complexity of the node-specific term, by keeping the position-specific component fixed (all three hierarchical levels included). 
We test the two different ways of distributing the shared embeddings, discussed in Section~\ref{sec:node-specific}. For each of these ways, we further test two cases to compute the node-specific term of the embeddings with respect to complexity and number of learnable parameters. We end up with the following four ways: 
\begin{enumerate}
    \item \textbf{Intra-partition shared embeddings:} The nodes that belong to the same partition in the coarser level of the hierarchy share $c=b/m_0$ embeddings. We use hashing for assigning the nodes to the shared embeddings. We try the following cases: (i) a single hash function, $h = 1$, combined with learnable node-specific weights - PosHashEmb Intra ($h=1$) method, (ii) two hash functions, $h = 2$, and learnable node-specific importance weights for the relative contribution of the $h$ component vectors coming from the $h$ hash functions - PosHashEmb Intra ($h=2$) method.  
    \item \textbf{Inter-partition shared embeddings:} We learn globally shared node embeddings, where we assign $b$ shared embeddings for the nodes, irrespective of the partition they belong to. Again, we test the two different ways mentioned above, which we call PosHashEmb Inter ($h=1$) and PosHashEmb Inter ($h=2$) methods, respectively. 
\end{enumerate}

\begin{table*}[!t]
  \caption{Comparison of the different ways to compute the node-specific component of the embedding.}
  \label{explore-node-part}
    \setlength{\tabcolsep}{0.1em}
  \begin{threeparttable}
      \begin{tabularx}{1.0\textwidth}{ 
        >{}X
        >{\centering\arraybackslash\hsize=0.7\hsize}X 
        >{\centering\arraybackslash\hsize=0.7\hsize}X 
        >{\centering\arraybackslash\hsize=0.7\hsize}X
        >{\centering\arraybackslash\hsize=0.7\hsize}X 
        >{\centering\arraybackslash\hsize=0.7\hsize}X 
        >{\centering\arraybackslash\hsize=0.7\hsize}X }
        \toprule
        & \multicolumn{2}{c}{\textbf{ogbn-arxiv}} & \multicolumn{2}{c}{\textbf{ogbn-products}} & \multicolumn{2}{c}{\textbf{ogbn-proteins}} \\
        Method & GCN     & GAT     & GRAPHSAGE  & GAT & \text{MWE-DGCN} & GAT \\
        \midrule
        \mbox{PosFullEmb} & \mbox{$0.684\pm0.002$} & \mbox{$0.680\pm0.001$} & \mbox{$0.762\pm0.004$} & \mbox{$0.758\pm0.003$} &\mbox{$0.782\pm0.015$} & \mbox{$0.757\pm0.006$}  \\
        PosHashEmb Inter $(h=1)$ & \mbox{$0.681\pm0.003$} & \mbox{$0.672\pm0.005$} & \mbox{$0.761\pm0.003$} & \mbox{$0.762\pm0.007$} & \mbox{$0.797\pm0.018$} &\mbox{$0.720\pm0.010$}   \\
        PosHashEmb Inter $(h=2)$ & \mbox{$0.681\pm0.002$} & \mbox{$0.668\pm0.004$} & \mbox{$0.760\pm0.003$} & \mbox{$0.767\pm0.009$} & \mbox{$0.792\pm0.021$} & \mbox{$0.721\pm0.014$}  \\
        PosHashEmb Intra $(h=1)$ & \mbox{$0.681\pm0.003$} & \mbox{$0.677\pm0.001$} & \mbox{$0.759\pm0.002$} & \mbox{$0.766\pm0.007$} &\mbox{$0.796\pm0.007$} & \mbox{$0.784\pm0.007$}  \\
        PosHashEmb Intra $(h=2)$ & \mbox{$0.683\pm0.001$} & \mbox{$0.671\pm0.004$} & \mbox{$0.756\pm0.004$} & \mbox{$0.764\pm0.004$} &\mbox{$0.786\pm0.022$} & \mbox{$0.787\pm 0.006$}  \\
        \bottomrule
      \end{tabularx}
        \begin{tablenotes}
            \item For all the methods, we keep fixed the position-specific component and we use all three hierarchical levels.
            PosFullEmb refers to a method which uses full embeddings for the computation of the node-specific component.
            For the Inter and Intra methods, the memory savings range from $88\%$ up to $97\%$ across all datasets and GNN models. 
            This corresponds to approximately $1/9$ and $1/34$ of the full size, respectively.
            Following OGB, the values for ogbn-arxiv and ogbn-products correspond to accuracy, and for ogbn-proteins to ROC-AUC metric; the higher the values the better. $h$ is the number of hash functions used. We set $\alpha=1/4$.
        \end{tablenotes}
  \end{threeparttable}
\end{table*}

We present the results in Table~\ref{explore-node-part}. As we can see, the performance is similar among the different ways to compute the node-specific component. In most cases, the performance is either the same or better compared to PosFullEmb, which induces memory requirements larger than the full size. This indicates that we do not need the full extent of the node-specific component for good performance. At the same time, the PosHashEmb Intra and Inter methods, achieve to reduce the amount of memory required to compute the initial node embeddings by a range from $88\%$ up to $97\%$ across all datasets and GNN models.

\subsection{Performance comparison with the baselines. (RQ5)}
We compare the performance of competing methods against our method PosHashEmb. The position-specific component of PosHashEmb is computed with PosEmb $3$-level and the node-specific with Intra $h=2$.
We test the performance when the memory requirements are approximately equal to $1/2, 1/6$ and $1/12$ of the full size for ogbn-arxiv and ogbn-proteins, and equal to $1/2, 1/18$ and $1/34$ of the full size for ogbn-products as there is much more room for memory reduction in the case of this larger dataset.  

For the hashing-based methods, the memory requirements are determined by the value of $B$. For PosHashEmb, we control the memory requirements by adjusting the value of $b$ of the node-specific component. When is needed, for the case of the smallest amount of memory, we use only the position-specific component for computing the node embeddings, e.g., PosEmb $1$-level with $k$ selected accordingly, in order to match the desired memory requirements. For DHE, we control the required amount of memory by adjusting the number of hidden layers and their width.
%
\renewcommand{\arraystretch}{0.7}
\begin{figure*}[!t]
    \centering
    \begin{tabular}{cc}
     \multicolumn{2}{c}{\includegraphics[width=140mm]{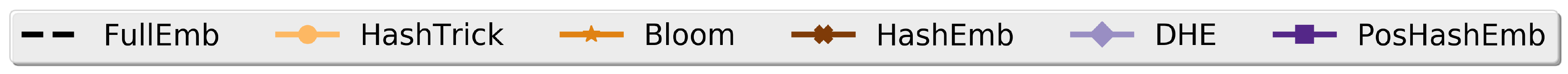}}\\
     \includegraphics[width=70mm]{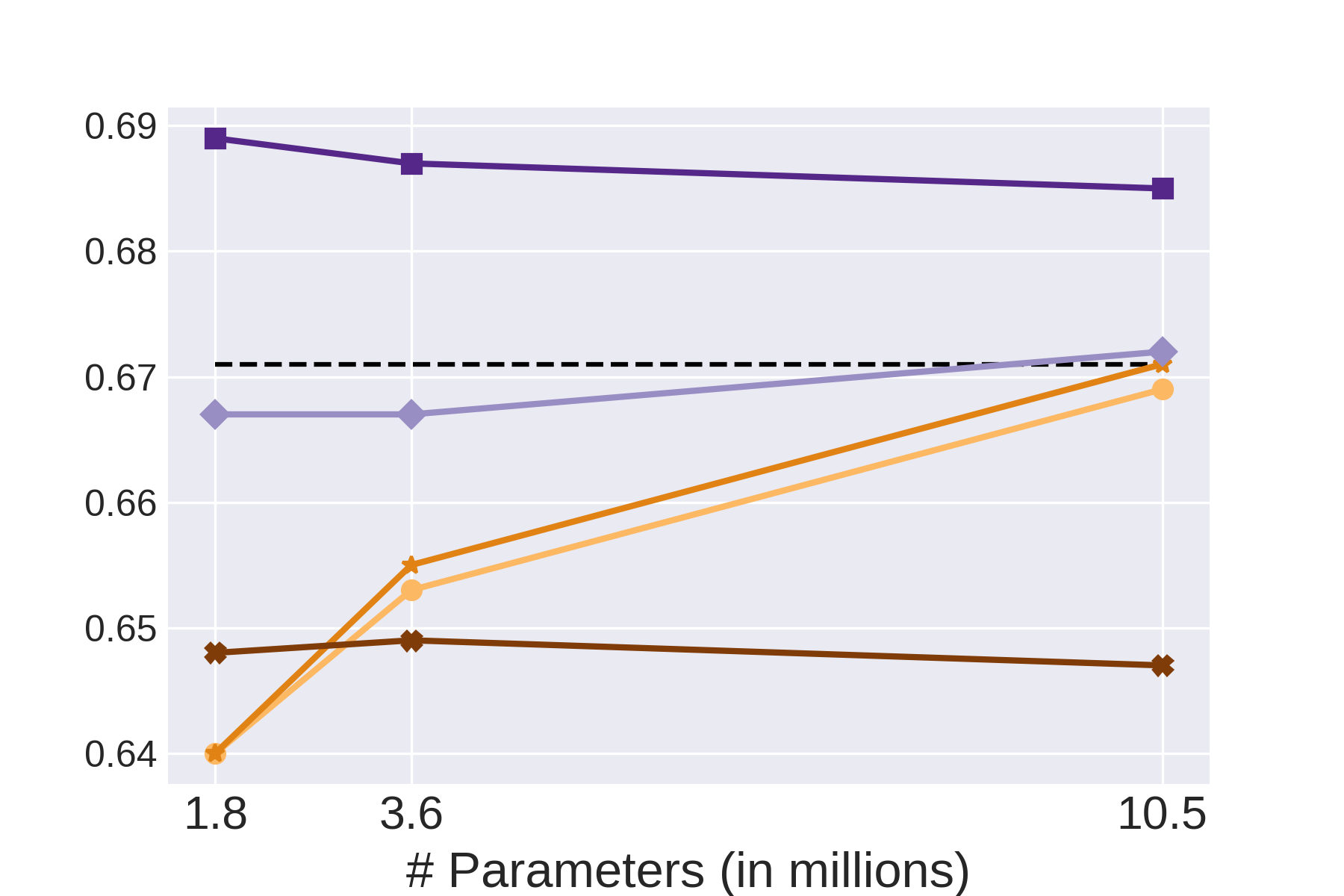} &
     \includegraphics[width=70mm]{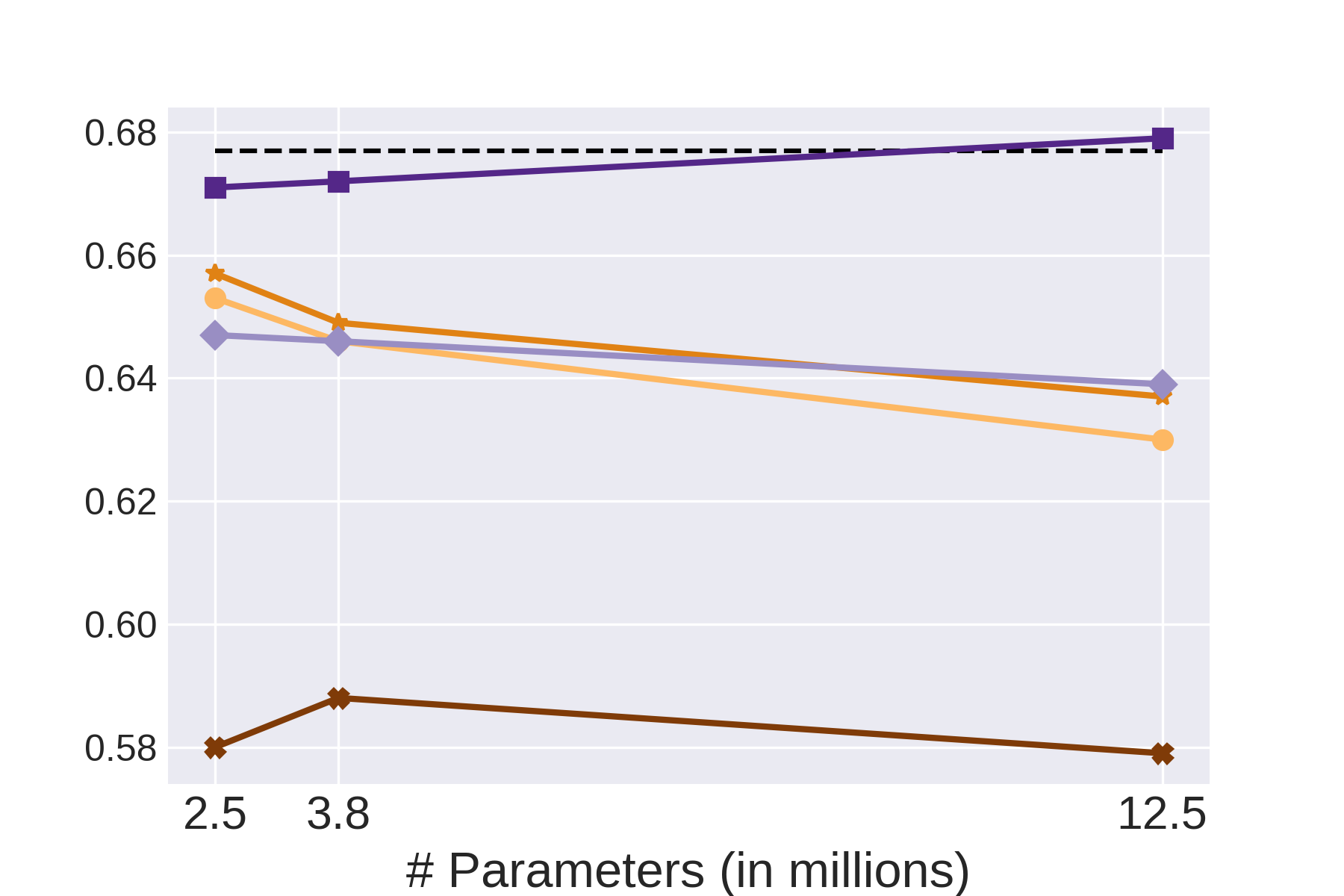} \\
     (a) ogbn-arxiv, GCN & (b) ogbn-arxiv, GAT \\
     \includegraphics[width=70mm]{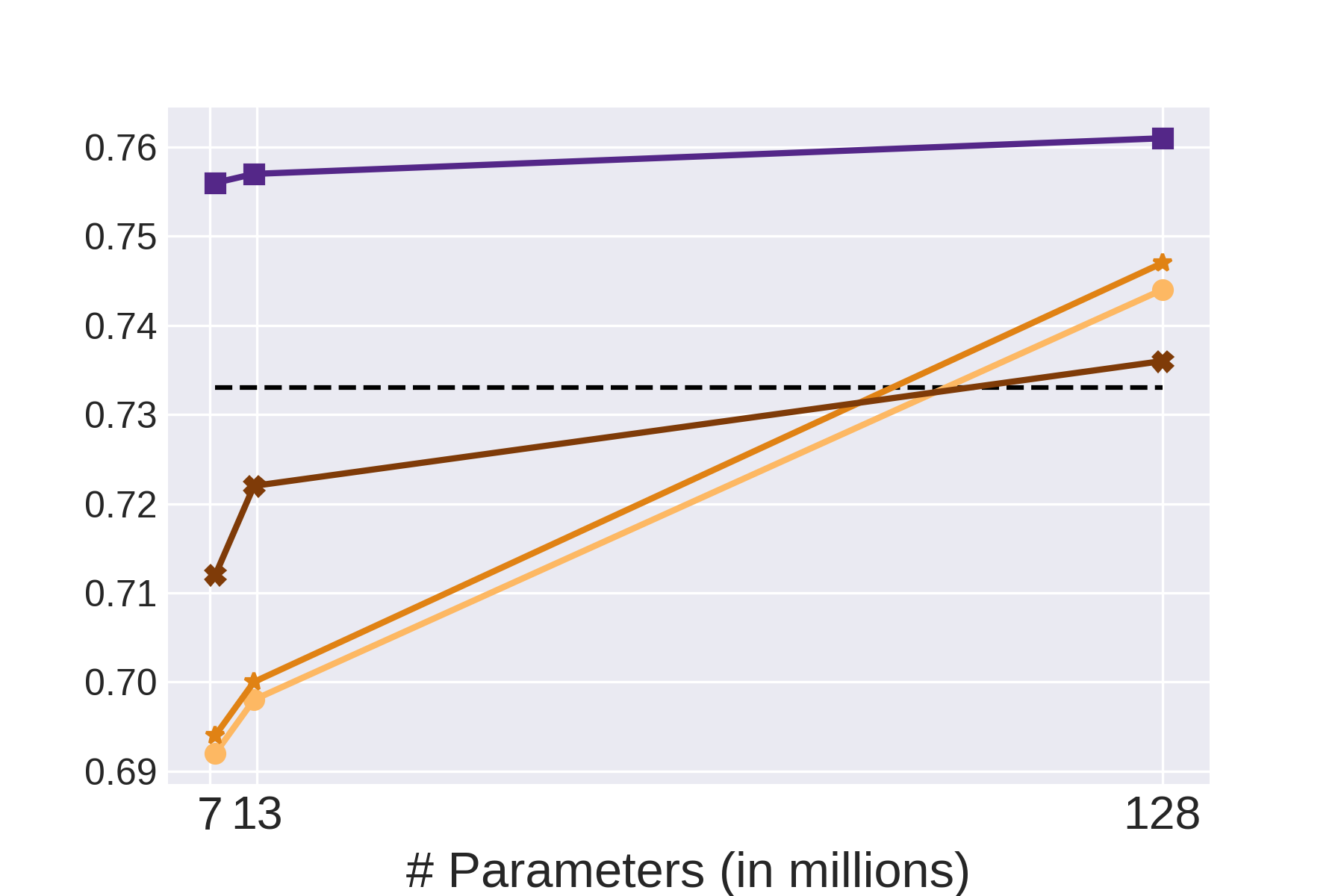} &
     \includegraphics[width=70mm]{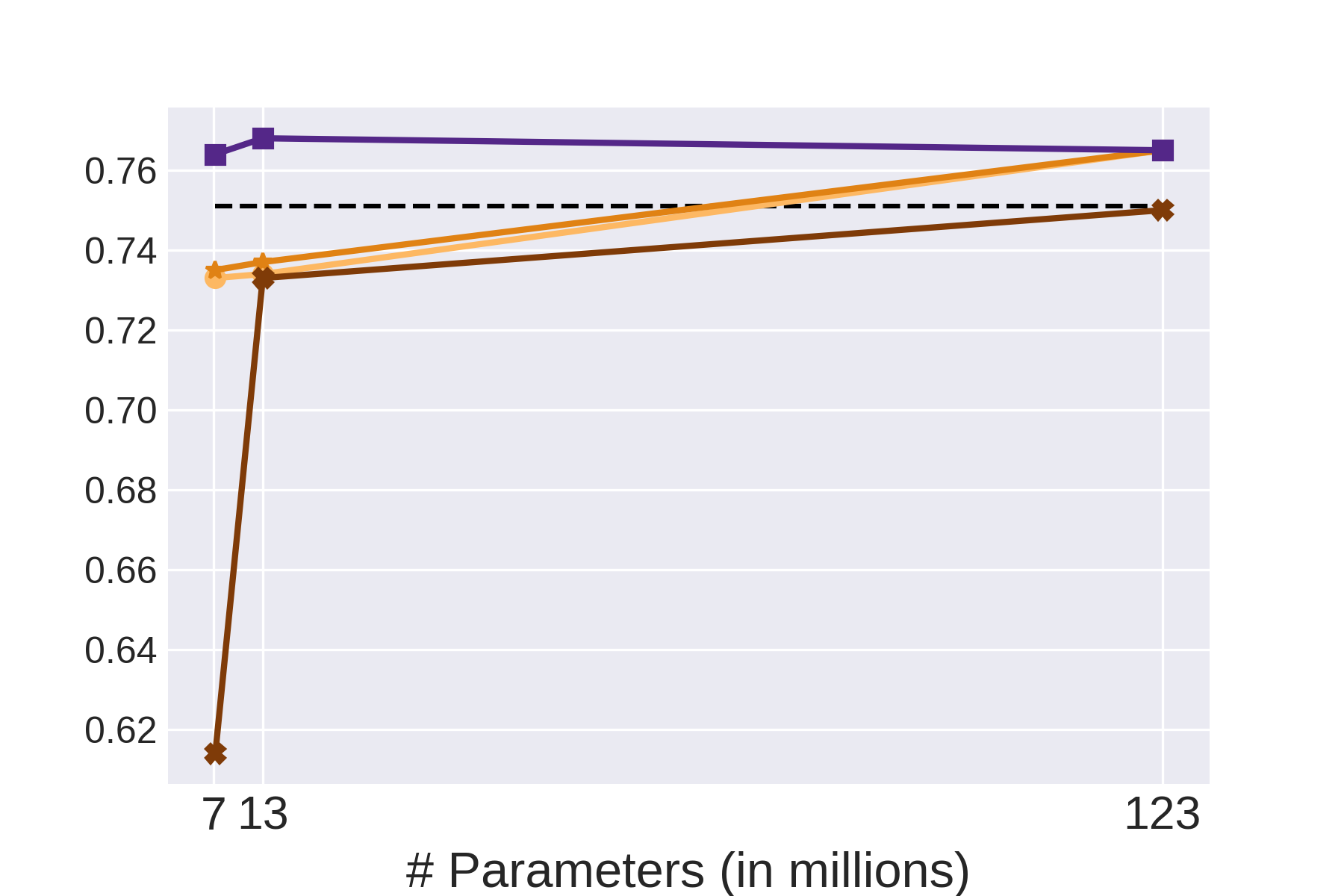} \\
     (c) ogbn-products, GRAPHSAGE & (d) ogbn-products, GAT \\
     \includegraphics[width=70mm]{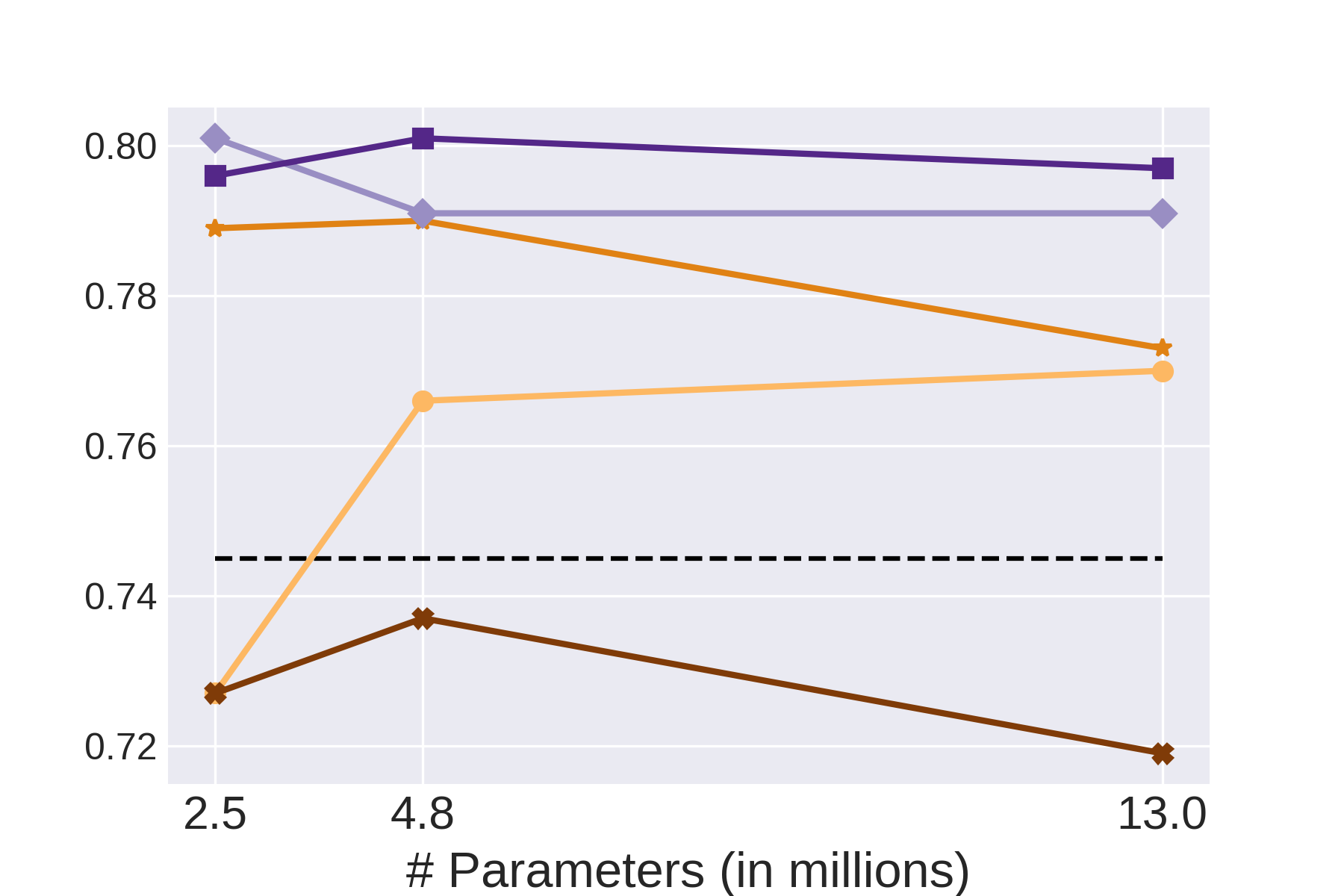} &
     \includegraphics[width=70mm]{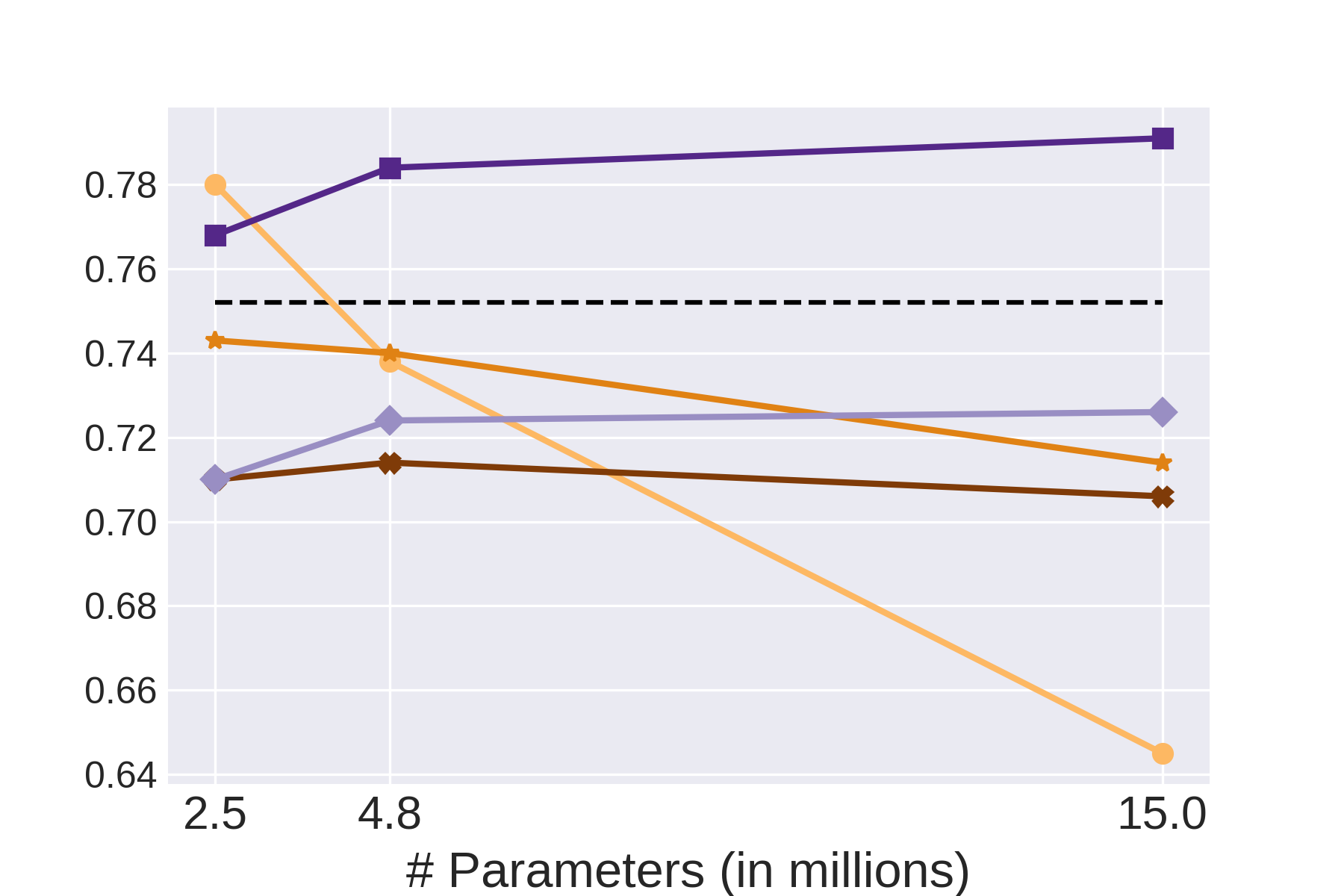} \\
     (e) ogbn-proteins, MWE-DGCN  & (f) ogbn-proteins, GAT  \\
    \end{tabular}
    \caption{Performance as a function of the amount of memory required to compute the initial node embeddings, in all datasets and GNN models. The number of trainable parameters of FullEmb for ogbn-arxiv is around 22M, for ogbn-products around 245M and for ogbn-proteins is around 28M. The plotted values for the number of parameters correspond to memory size which is approximately equal to $1/12, 1/6$ and $1/2$ of the full size for ogbn-arxiv and ogbn-proteins, and equal to $1/34, 1/18$ and $1/2$ of the full size for ogbn-products. 
    PosHashEmb uses PosEmb $3$-level for computing the position-specific component and Intra $h=2$ for the node-specific component. Following OGB, for ogbn-arxiv and ogbn-products the performance is measured with accuracy, and for ogbn-proteins with ROC-AUC metric; the higher the values the better.}
    \label{fig:baselines}
\end{figure*}
%
For the largest dataset ogbn-products, we were not able to run DHE with the same batch size of $1000$ as all other methods, because of GPU memory limitations. 
While trying smaller batch sizes of $500, 300, 100$, we observed that the performance significantly deteriorated and thus, we do not report these results.

The results are presented in Figure~\ref{fig:baselines}. As we can see, PosHashEmb performs better than FullEmb in all cases except for ogbn-arxiv and GAT. This is true even in the case of the smallest number of trainable parameters. According to our results, PosHashEmb performs better than all other approaches; this is true even in the cases where we have reduced the number of trainable parameters the most, with the exception of the ogbn-proteins dataset. Note that for the largest dataset, ogbn-products, PosHashEmb has the best performance and at the same time achieves up to $35$ times less parameters or $97\%$ memory savings compared to full embeddings.  
Another important observation is that the performance of PosHashEmb does not seem to vary significantly as we modify the amount of memory required to compute the initial node embeddings across all datasets and GNN models. This indicates that we are able to achieve whatever memory savings required based on the available resources, while keeping the performance high.

\section{Related work}
\label{rw}
Hashing has been widely used for compression of the feature space in recommender systems and NLP applications. Hashing trick~\cite{weinberger2009feature} is the simplest technique that uniformly maps feature values to a smaller number of shared hash buckets. 
Serra et al.~\cite{serra2017getting} motivated by bloom filters~\cite{bloom1970space}, they use multiple hash functions to generate binary encodings. Then a linear layer is applied to the encoding to recover the embedding for the given feature value.
Hash embeddings~\cite{svenstrup2017hash} use multiple hash functions to retrieve multiple entries from the shared embedding table and then combine them to generate the final embeddings.
The main contribution of this method compared to others using multiple hash functions, is that it learns importance weights dedicated to each feature value which control the contribution of each entry to the final embedding.

In recommender systems, Zhang et al.~\cite{zhang2020model} separate the features based on their frequency, as an indicator of their importance. They ensure that the most frequent ones will be assigned a unique embedding (zero collisions for those) and the rest will be hashed to shared embeddings using two hash functions (double hashing).  
In a recent work, Shi et al.~\cite{shi2020compositional} create a unique embedding for each category by composing shared entries from multiple smaller embedding tables.
Again in the recommendation domain, Kang et al.~\cite{kang2020deep} propose DHE that replaces one-hot encodings with dense vectors from multiple hash functions. Then it trains a feedforward network to produce the final embeddings.

While all the aforementioned methods are designed to reduce or eliminate collisions coming from hashing, there is another line of work, similar to ours, that uses hashing as a means to maintain a notion of similarity. 
Locality-sensitive hashing~\cite{datar2004locality} hashes similar features into the same buckets and as such, aims to maximize collisions.
HashRec~\cite{kang2019candidate} is a learning-to-hash method that learns preference-preserving binary codes for users and items in top-$N$ recommendation. By using the hamming distance, it estimates preferences between users and items.



\section{Conclusion}
\label{conclusion}
In this work, we study the problem of node classification using GNNs and we focus on the case when there are no available node features or the features are weak. In such cases, GNNs use one-hot encodings to learn an embedding layer and compute node embeddings that can then be used as input features; this can be memory expensive and is not scalable for large graphs. We present a family of methods that take into account the nodes' position in the graph to compute efficient node embeddings and reduce the number of trainable parameters significantly. 
Our final embeddings are generated by the combination of a position-specific component and a node-specific component; the so-called PosHashEmb method.
The starting point for all our methods is to discover communities of similar nodes. For each partition we learn a unique embedding. Then, we build up the complexity of the embeddings by creating multiple hierarchical layers of communities and by including a node-specific term. For the latter, we use hashing and node-dedicated learnable weights to enhance the embedding with more localized, finer, node-specific signals. 

The complexity of the node-specific component can vary depending on the desired model compression. It could even be excluded entirely when we need to induce extremely high model compression. In such cases, only the position-specific component is used and as we showed, PosEmb $3$-level has better or similar performance with full embeddings, and at the same time achieves to reduce the amount of memory required to compute the initial node embeddings by $90\%$ up to $99\%$ across all datasets and GNN models.
Our methods model homophily which is a strong characteristic of many real-world graphs, and as our experimental results showed, this is highly beneficial for the performance. Specifically, our method PosHashEmb, that combines both components, performs better than both state-of-the-art hash-based techniques and full embeddings in almost all cases, and at the same time achieves great memory savings, reaching $88\%$ up to $97\%$ of the full size across the different datasets and GNN models.  



\section*{Acknowledgment}

This work was supported in part by NSF (1447788, 1704074, 1757916, 1834251), Army Research Office (W911NF1810344), and the Digital Technology Center at the University of Minnesota. Access to research and computing facilities was provided by the Digital Technology Center and the Minnesota Supercomputing Institute.


\bibliographystyle{IEEEtran}
\bibliography{IEEEabrv,mybib}

\begin{thebibliography}{10}
\providecommand{\url}[1]{#1}
\csname url@samestyle\endcsname
\providecommand{\newblock}{\relax}
\providecommand{\bibinfo}[2]{#2}
\providecommand{\BIBentrySTDinterwordspacing}{\spaceskip=0pt\relax}
\providecommand{\BIBentryALTinterwordstretchfactor}{4}
\providecommand{\BIBentryALTinterwordspacing}{\spaceskip=\fontdimen2\font plus
\BIBentryALTinterwordstretchfactor\fontdimen3\font minus
  \fontdimen4\font\relax}
\providecommand{\BIBforeignlanguage}[2]{{%
\expandafter\ifx\csname l@#1\endcsname\relax
\typeout{** WARNING: IEEEtran.bst: No hyphenation pattern has been}%
\typeout{** loaded for the language `#1'. Using the pattern for}%
\typeout{** the default language instead.}%
\else
\language=\csname l@#1\endcsname
\fi
#2}}
\providecommand{\BIBdecl}{\relax}
\BIBdecl

\bibitem{te2018rgcnn}
G.~Te, W.~Hu, A.~Zheng, and Z.~Guo, ``Rgcnn: Regularized graph cnn for point
  cloud segmentation,'' in \emph{Proceedings of the 26th ACM international
  conference on Multimedia}, 2018, pp. 746--754.

\bibitem{marcheggiani2018exploiting}
D.~Marcheggiani, J.~Bastings, and I.~Titov, ``Exploiting semantics in neural
  machine translation with graph convolutional networks,'' \emph{arXiv preprint
  arXiv:1804.08313}, 2018.

\bibitem{manchanda2021importance}
S.~Manchanda and G.~Karypis, ``Importance assessment in scholarly networks,''
  in \emph{CEUR Workshop Proceedings}, vol. 2831.\hskip 1em plus 0.5em minus
  0.4em\relax CEUR-WS, 2021.

\bibitem{shui2020heterogeneous}
Z.~Shui and G.~Karypis, ``Heterogeneous molecular graph neural networks for
  predicting molecule properties,'' \emph{arXiv preprint arXiv:2009.12710},
  2020.

\bibitem{ying2018graph}
R.~Ying, R.~He, K.~Chen, P.~Eksombatchai, W.~L. Hamilton, and J.~Leskovec,
  ``Graph convolutional neural networks for web-scale recommender systems,'' in
  \emph{Proceedings of the 24th ACM SIGKDD International Conference on
  Knowledge Discovery \& Data Mining}, 2018, pp. 974--983.

\bibitem{weinberger2009feature}
K.~Weinberger, A.~Dasgupta, J.~Langford, A.~Smola, and J.~Attenberg, ``Feature
  hashing for large scale multitask learning,'' in \emph{Proceedings of the
  26th annual international conference on machine learning}, 2009, pp.
  1113--1120.

\bibitem{svenstrup2017hash}
D.~Svenstrup, J.~M. Hansen, and O.~Winther, ``Hash embeddings for efficient
  word representations,'' \emph{arXiv preprint arXiv:1709.03933}, 2017.

\bibitem{kang2020deep}
W.-C. Kang, D.~Z. Cheng, T.~Yao, X.~Yi, T.~Chen, L.~Hong, and E.~H. Chi, ``Deep
  hash embedding for large-vocab categorical feature representations,''
  \emph{arXiv preprint arXiv:2010.10784}, 2020.

\bibitem{serra2017getting}
J.~Serra and A.~Karatzoglou, ``Getting deep recommenders fit: Bloom embeddings
  for sparse binary input/output networks,'' in \emph{Proceedings of the
  Eleventh ACM Conference on Recommender Systems}, 2017, pp. 279--287.

\bibitem{zhang2020model}
C.~Zhang, Y.~Liu, Y.~Xie, S.~I. Ktena, A.~Tejani, A.~Gupta, P.~K. Myana,
  D.~Dilipkumar, S.~Paul, I.~Ihara \emph{et~al.}, ``Model size reduction using
  frequency based double hashing for recommender systems,'' in \emph{Fourteenth
  ACM Conference on Recommender Systems}, 2020, pp. 521--526.

\bibitem{mcpherson2001birds}
M.~McPherson, L.~Smith-Lovin, and J.~M. Cook, ``Birds of a feather: Homophily
  in social networks,'' \emph{Annual review of sociology}, vol.~27, no.~1, pp.
  415--444, 2001.

\bibitem{hu2020open}
W.~Hu, M.~Fey, M.~Zitnik, Y.~Dong, H.~Ren, B.~Liu, M.~Catasta, and J.~Leskovec,
  ``Open graph benchmark: Datasets for machine learning on graphs,''
  \emph{arXiv preprint arXiv:2005.00687}, 2020.

\bibitem{carter1979universal}
J.~L. Carter and M.~N. Wegman, ``Universal classes of hash functions,''
  \emph{Journal of computer and system sciences}, vol.~18, no.~2, pp. 143--154,
  1979.

\bibitem{bloom1970space}
B.~H. Bloom, ``Space/time trade-offs in hash coding with allowable errors,''
  \emph{Communications of the ACM}, vol.~13, no.~7, pp. 422--426, 1970.

\bibitem{wang2019dgl}
M.~Wang, D.~Zheng, Z.~Ye, Q.~Gan, M.~Li, X.~Song, J.~Zhou, C.~Ma, L.~Yu,
  Y.~Gai, T.~Xiao, T.~He, G.~Karypis, J.~Li, and Z.~Zhang, ``Deep graph
  library: A graph-centric, highly-performant package for graph neural
  networks,'' \emph{arXiv preprint arXiv:1909.01315}, 2019.

\bibitem{kipf2016semi}
T.~N. Kipf and M.~Welling, ``Semi-supervised classification with graph
  convolutional networks,'' \emph{arXiv preprint arXiv:1609.02907}, 2016.

\bibitem{velivckovic2017graph}
P.~Veli{\v{c}}kovi{\'c}, G.~Cucurull, A.~Casanova, A.~Romero, P.~Lio, and
  Y.~Bengio, ``Graph attention networks,'' \emph{arXiv preprint
  arXiv:1710.10903}, 2017.

\bibitem{hamilton2017inductive}
W.~L. Hamilton, R.~Ying, and J.~Leskovec, ``Inductive representation learning
  on large graphs,'' \emph{arXiv preprint arXiv:1706.02216}, 2017.

\bibitem{karypis1997metis}
G.~Karypis and V.~Kumar, ``Metis: A software package for partitioning
  unstructured graphs, partitioning meshes, and computing fill-reducing
  orderings of sparse matrices,'' 1997.

\bibitem{ioffe2015batch}
S.~Ioffe and C.~Szegedy, ``Batch normalization: Accelerating deep network
  training by reducing internal covariate shift,'' in \emph{International
  conference on machine learning}.\hskip 1em plus 0.5em minus 0.4em\relax PMLR,
  2015, pp. 448--456.

\bibitem{shi2020compositional}
H.-J.~M. Shi, D.~Mudigere, M.~Naumov, and J.~Yang, ``Compositional embeddings
  using complementary partitions for memory-efficient recommendation systems,''
  in \emph{Proceedings of the 26th ACM SIGKDD International Conference on
  Knowledge Discovery \& Data Mining}, 2020, pp. 165--175.

\bibitem{datar2004locality}
M.~Datar, N.~Immorlica, P.~Indyk, and V.~S. Mirrokni, ``Locality-sensitive
  hashing scheme based on p-stable distributions,'' in \emph{Proceedings of the
  twentieth annual symposium on Computational geometry}, 2004, pp. 253--262.

\bibitem{kang2019candidate}
W.-C. Kang and J.~McAuley, ``Candidate generation with binary codes for
  large-scale top-n recommendation,'' in \emph{Proceedings of the 28th ACM
  International Conference on Information and Knowledge Management}, 2019, pp.
  1523--1532.

\end{thebibliography}

\end{document}